\documentclass[10pt,twocolumn,letterpaper]{article}

\usepackage[final]{cvpr}              
\pdfoutput=1
\usepackage{graphicx}
\usepackage{amsmath}
\usepackage{amssymb}
\usepackage{amsfonts}
\usepackage{nicefrac}       
\usepackage{mathrsfs}
\usepackage{mathtools}
\usepackage{tabularx}
\usepackage{array}          
\usepackage{booktabs}
\usepackage{longtable}
\usepackage{wrapfig}
\usepackage{rotating}
\usepackage{multirow}
\usepackage{caption}
\usepackage{lipsum}
\usepackage[dvipsnames]{xcolor}
\usepackage{enumitem}
\usepackage{inconsolata}
\usepackage{pifont}
\usepackage{soul}
\usepackage{siunitx}
\usepackage{csquotes}
\usepackage{xspace}
\usepackage{comment}
\usepackage{adjustbox}
\usepackage{algorithm}
\usepackage{algpseudocode}
\usepackage{fontawesome}
\usepackage{marvosym}
\usepackage{pifont}
\usepackage{colortbl}
\usepackage{arydshln}
\usepackage[framemethod=TikZ]{mdframed}

\usepackage[normalem]{ulem}
\usepackage[accsupp]{axessibility}  

\definecolor{cvprblue}{rgb}{0.21,0.49,0.74}
\usepackage[pagebackref,breaklinks,colorlinks,allcolors=cvprblue]{hyperref}

\newcommand{\velocitilogo}[1]{\includegraphics[scale={#1}]{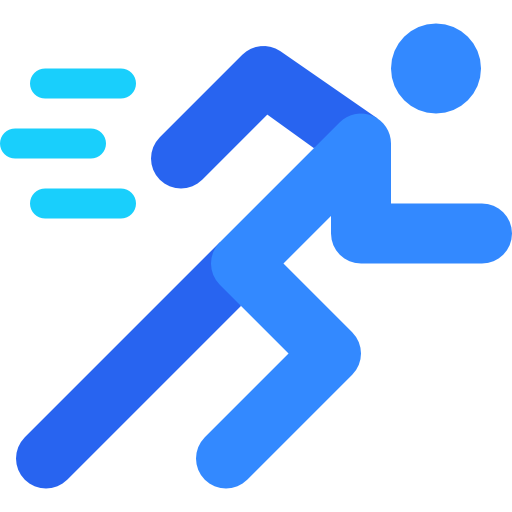}}
\newcommand{\tabpub}[1]{{\scriptsize \color{gray}{#1}}}
\newcommand{\NA}{{\scriptsize \color{gray}\textit{NA}}}

\definecolor{velopink}{RGB}{190, 60, 108}

\definecolor{veloblue}{RGB}{83, 149, 218}
\definecolor{veloorange}{RGB}{236, 133, 45}

\definecolor{actioncolor}{RGB}{79, 173, 91}
\definecolor{agentcolor}{RGB}{104, 51, 154}
\definecolor{meucolor}{RGB}{222, 48, 135}

\newcommand*\myglobe{%
\includegraphics[height=1.6ex]{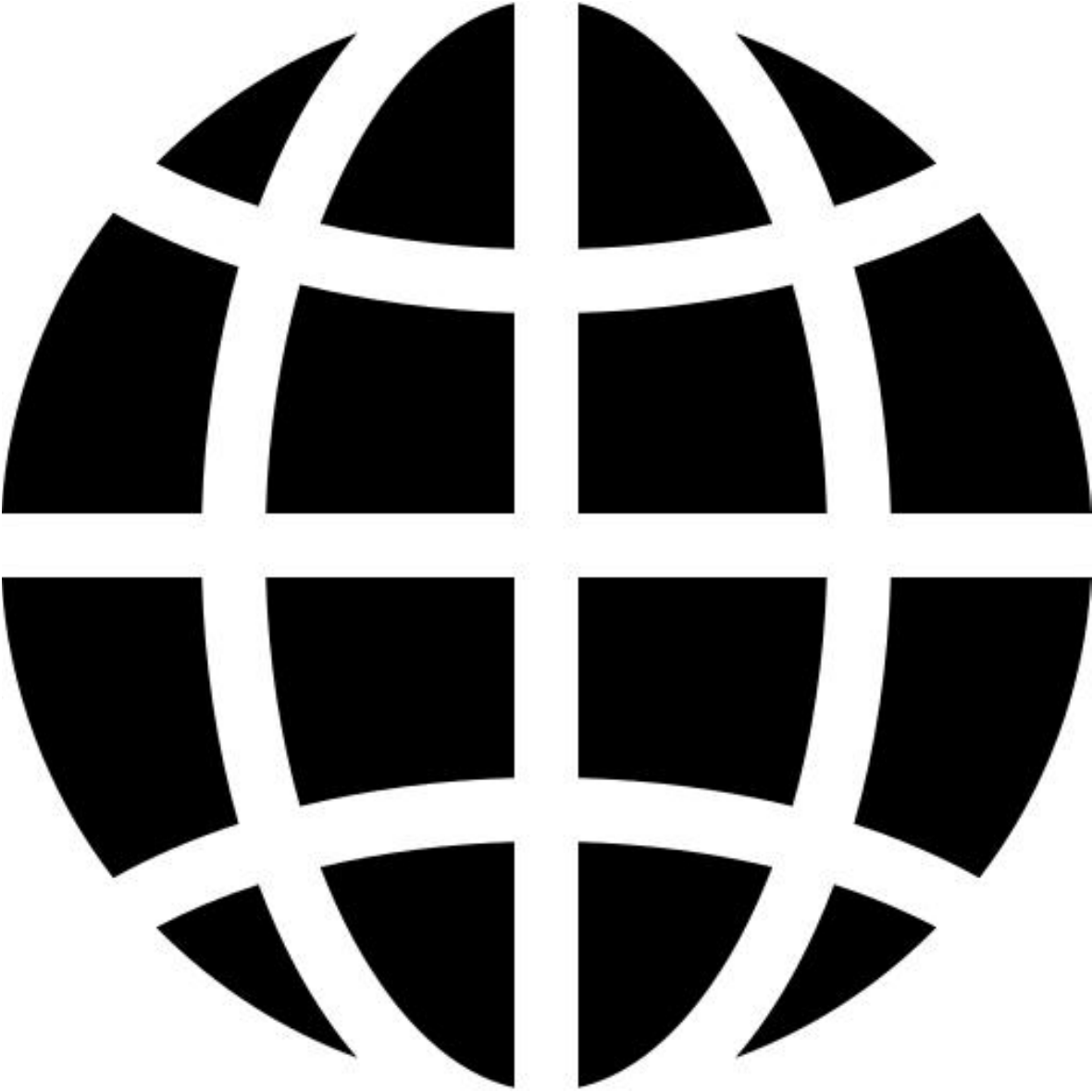}}

\makeatletter
\def\thanks#1{\protected@xdef\@thanks{\@thanks
        \protect\footnotetext{#1}}}
\makeatother

\definecolor{strictvlecolor}{RGB}{246, 246, 210}
\definecolor{scatcorrect}{RGB}{166, 227, 136}
\definecolor{scatstrict}{RGB}{46, 77, 27}
\definecolor{scatincorrect}{RGB}{203, 59, 39}

\definecolor{qablue}{RGB}{190, 214, 234}
\newcommand{\qabluetext}[1]{{\color{qablue}#1}}

\newcommand{\bluetext}[1]{{\color{veloblue}#1}}
\newcommand{\orangetext}[1]{{\color{veloorange}#1}}

\newcommand{\agenttext}[1]{{\color{agentcolor}#1}}
\newcommand{\actiontext}[1]{{\color{actioncolor}#1}}
\newcommand{\meutext}[1]{{\color{meucolor}#1}}

\definecolor{forestgreen}{rgb}{0.13, 0.55, 0.13}
\definecolor{fireenginered}{rgb}{0.81, 0.09, 0.13}

\newcommand{\cmark}{{\textcolor{OliveGreen}{\ding{51}}}}%
\newcommand{\xmark}{{\textcolor{red}{\ding{55}}}}%

\renewcommand{\emph}[1]{\textit{#1}}
\renewcommand{\paragraph}[1]{\vspace{0.8mm}\noindent\textbf{#1}}

\newcommand{\benchmark}{VELOCITI}

\newcommand{\control}{Control Test}
\newcommand{\argen}{Agent Random Test}
\newcommand{\arghn}{Agent Binding Test}
\newcommand{\coref}{Agent Coreference Test}
\newcommand{\verbgen}{Action Adversarial Test}
\newcommand{\verbret}{Action Binding Test}
\newcommand{\manner}{Action Manner Test}
\newcommand{\order}{Event Chronology Test}

\newcommand{\argensh}{AgRand}
\newcommand{\arghnsh}{AgBind}
\newcommand{\corefsh}{AgCref}
\newcommand{\verbgensh}{ActAdv}
\newcommand{\verbretsh}{ActBind}
\newcommand{\mannersh}{ActMan}
\newcommand{\ordersh}{EvChr}

\newcommand{\strictvle}{StrictVLE}
\newcommand{\classicvle}{ClassicVLE}
\newcommand{\quoteyes}{\lq{}Yes\rq{}}
\newcommand{\quoteno}{\lq{}No\rq{}}

\newcommand{\llm}{LLM}
\newcommand{\vllm}{Video-LLM}

\newcommand{\ovseven}{LLaVA-OneVision-7B}
\newcommand{\ovsevensi}{LLaVA-OneVision-7B-Si}

\newcommand{\ovseventy}{LLaVA-OneVision-72B}
\newcommand{\pllava}{P-LLaVA}
\newcommand{\videocon}{Owl-Con}
\newcommand{\vidllava}{Video-LLaVA}
\newcommand{\geminiflash}{Gemini-1.5-Flash}
\newcommand{\gpt}{GPT-4o}
\newcommand{\qwenvl}{Qwen2-VL}

\newcommand{\ov}{LLaVA-OneVision}

\newcommand{\qwenvlseventy}{Qwen2-VL-72B}

\def\paperID{8830} 
\def\confName{CVPR}
\def\confYear{2025}

\title{\velocitilogo{0.035} VELOCITI: Benchmarking Video-Language Compositional Reasoning \\ with Strict Entailment
\vspace{-3mm}
}

\author{
Darshana Saravanan$^{1\textsuperscript{\Yinyang}}$ \hspace{1cm}
Varun Gupta$^{1\textsuperscript{\Yinyang}}$ \hspace{1cm}
Darshan Singh$^{1\textsuperscript{\Yinyang}}$ \hspace{1cm}
Zeeshan Khan$^2$ \\
Vineet Gandhi$^1$ \hspace{1cm}
Makarand Tapaswi$^1$ \vspace{1mm} \\
{\normalsize $^1$CVIT, IIIT Hyderabad, India}%
\hspace{1cm} {\normalsize $^2$Inria, Paris, France} \\
{\texttt{\normalsize ~\myglobe~\href{https://katha-ai.github.io/projects/velociti}{katha-ai.github.io/projects/velociti}}}
\vspace{-2mm}
\thanks{\Yinyang~~Equal contribution}
}

\begin{document}
\maketitle
\begin{abstract}

A fundamental aspect of compositional reasoning in a video is associating people and their actions across time.
Recent years have seen great progress in general-purpose vision/video models and a move towards long-video understanding.
While exciting, we take a step back and ask: \textit{are today's models good at compositional reasoning on short videos?}
To this end, we introduce \benchmark{}, a benchmark to study \vllm{}s by disentangling and assessing the comprehension of agents, actions, and their associations across multiple events.
We adopt the Video-Language Entailment setup and propose \strictvle{} that requires correct classification (rather than ranking) of the positive and negative caption.
We evaluate several models and observe that even the best, LLaVA-OneVision (44.5\%) and Gemini-1.5-Pro (49.3\%), are far from human accuracy at 93.0\%.
Results show that action understanding lags behind agents, and negative captions created using entities appearing in the video perform worse than those obtained from pure text manipulation.
We also present challenges with \classicvle{} and multiple-choice (MC) evaluation, strengthening our preference for \strictvle{}.
Finally, we validate that our benchmark requires visual inputs of multiple frames making it ideal to study video-language compositional reasoning.
\end{abstract}
\vspace{-2mm}
\section{Introduction}
\label{sec:intro}

\begin{figure}[t]
\centering
\includegraphics[width=0.8\linewidth]{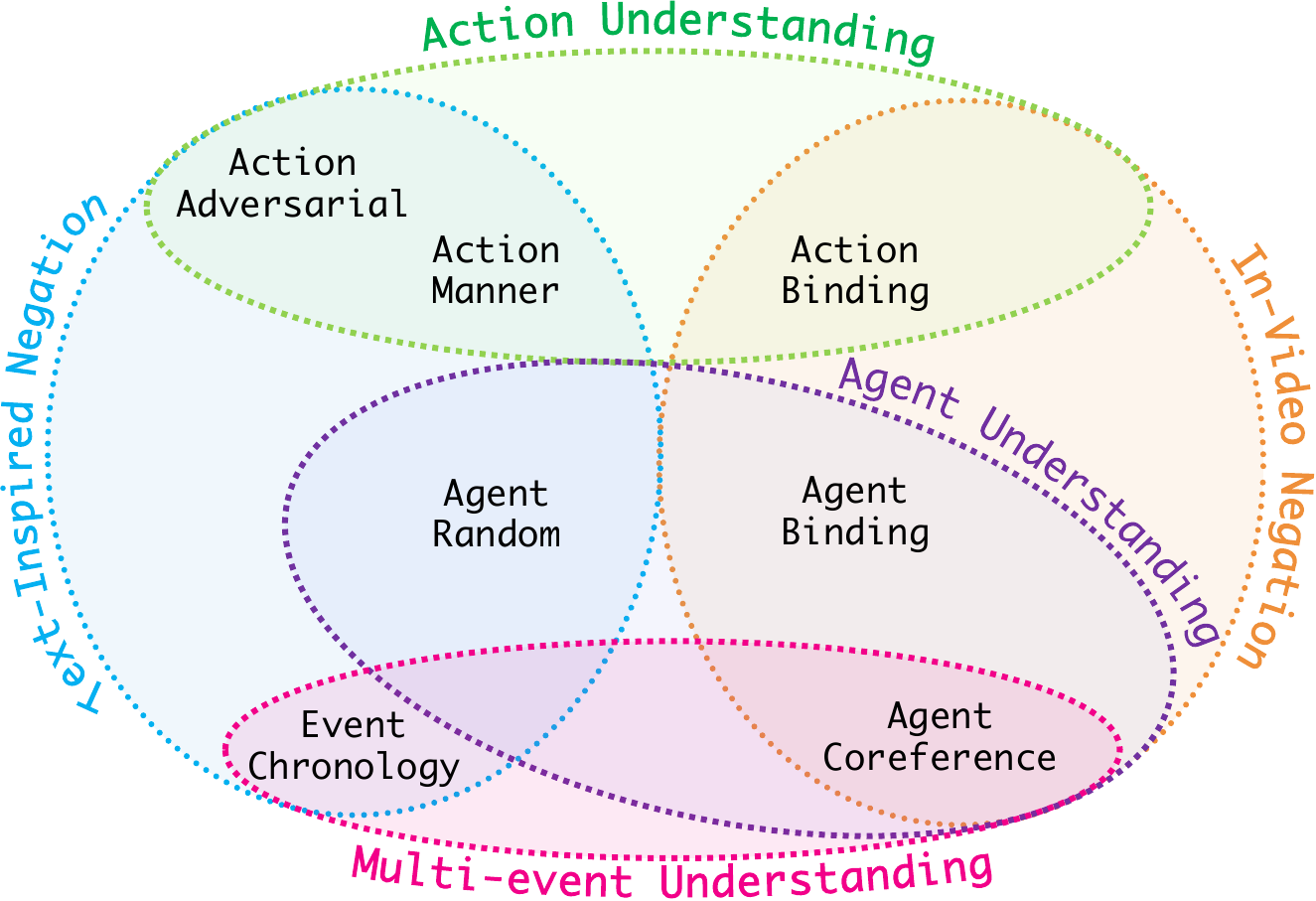}
\vspace{-1mm}
\caption{A Venn diagram grouping \benchmark{}'s seven tests (in black) that evaluate a \vllm{} across different facets:
\agenttext{Agent Understanding},
\actiontext{Action Understanding}, and
\meutext{Multi-event Understanding}.
The benchmark is formulated as video-language entailment, where negative captions are created by manipulating text
(\bluetext{Text-inspired Negation}) or from other parts of the same video (\orangetext{In-Video Negation}).
Best seen in color.
}
\label{fig:velociti_venn}
\vspace{-5mm}
\end{figure}

\emph{Near a parking lot, a man in a black hat smiles in a friendly way at a woman in a purple shirt.}
To a reader, this dense description paints a clear picture about a short snippet (event) of a video clip.
We build a mental model of two people (referred here by their clothing), at a specified location, and a short interaction between them.
Reading further,
\emph{the woman claps as a man in grey pants spins on one leg}.
We are able to associate that it is the same woman who is now cheering at a third person (likely) that is performing stunts.

The above example illustrates an intelligent agent's ability to perform compositional reasoning.
For video-language models, we scope this in two steps:
(i)~comprehend atomic entities, \eg~\textit{people} and \textit{actions}; and
(ii)~reason about them compositionally and across time by building associations%
\footnote{Associations can be thought as \textit{implicit} tuples that a model attempts to build while watching a video. Some examples include
\textit{person-attribute} tuples:
(man1, black hat),
(woman, purple shirt),
(man2, grey pants);
\textit{agent-action} tuples:
(man1, smiles at, woman),
(man2, spins on one leg),
(woman, claps at, man2);
or \textit{action-manner} tuples:
(smile, friendly way).
}.

In recent years, strong visual (image) encoders are combined with powerful Large Language Models (LLMs) to advance general-purpose vision~\cite{instructblip, qwen2VL, onevision, molmo, paligemma}.
A similar approach is adapted for videos to create \vllm{}s~\cite{pllava, videollava, qwen2VL, onevision}.
Keeping pace with the development of new models, there is a flurry of work on evaluating them (\cref{tab:BenchmarkComparison}).
Video researchers are also creating benchmarks to study long video comprehension~\cite{cinepile, mmbenchvideo, videomme, temporalbench}.
However, we take a step back and ask, \textit{are today's \vllm{}s ready to take on such challenges?}
Specifically, are they good at compositional reasoning in short videos, arguably a prerequisite to tackle complex and long videos?

To this end, we introduce the \textit{\benchmark{}}, a benchmark that studies
\textbf{V}ideo \textbf{e}t \textbf{L}anguage \textbf{C}ompos\textbf{i}tionality through \textbf{Ti}me.
We adopt the video-language entailment (VLE) evaluation setup~\cite{videocon} where a model is prompted to predict whether a video entails a caption 
(\quoteyes{} for an aligned or positive caption and \quoteno{} for a misaligned or negative caption).
Through a suite of seven tests, we are able to disentangle and assess a model's ability to comprehend \textit{agents}, \textit{actions}, and their associations across \textit{multiple events} through time.
As illustrated in \cref{fig:velociti_venn}, we group the 7 tests based on:
(i)~the specific facet of a model's ability (agent, action, multi-event), or
(ii)~the strategy used to create the negative caption (\bluetext{Text-Inspired Negation} \vs~\orangetext{In-Video Negation}).
Note that although the tests (\cref{subsec:tests}) have varying levels of difficulty,
they are all important as they shed light on whether a model is able to solve a specific facet of video-language compositional reasoning.

Our videos are sourced from the VidSitu dataset and are accompanied by action and semantic role label (SRL) annotations for multiple events in a short movie clip~\cite{vidsitu}.
The videos are diverse and feature multiple agents and actions across complex editing and shot changes,
while dense SRL succinctly describes \textit{who} did \textit{what} with/to \textit{whom}, \textit{where}, and (sometimes) \textit{how}.
Importantly, each SRL only describes a single event, requiring models to implicitly localize the event in the video before solving the test.

\paragraph{Strict entailment.}
In the classic VLE setup, benchmarks typically check if the entailment score for the positive caption is higher than the negative caption~\cite{sanyalmachines, visualentailmentimages, vitatecs}.
While this traces back to visual-semantic embedding models~\cite{devise, vseplusplus, clip}, it is unsuitable for evaluating modern \vllm{}s that generate text (and not similarity scores).

We propose a strict entailment scoring mechanism where \vllm{}s should output \quoteyes{} for an aligned caption and \quoteno{} for a misaligned one.
Our analysis reveals that models produce marginally different entailment scores for the positive and negative captions attaining good performance on \classicvle{}, but predict \quoteyes{} for both.
This is critical as VLE evaluates a model's ability to reject (partially) misaligned descriptions.
Poor performance here implies that the model may produce erroneous outputs on other tasks (\eg~question-answering) and assumes that partial hallucinations (like negative captions) are acceptable.

\paragraph{Contributions.}
We summarize our contributions and findings below:
(i)~We propose \benchmark{}, a new benchmark that evaluates compositional reasoning of video-language models.
Our test suite sheds light on a model's ability to perceive and reason about \textit{agents} and \textit{actions} across \textit{multiple events}, identifying challenges for improvement
(\cref{sec:benchmark}).
(ii)~We propose a strict metric for video-language entailment
that requires a model to produce \quoteyes{} for an aligned caption \textit{and} \quoteno{} for the corresponding misaligned caption
(\cref{sec:strictvle_metric}).
(iii)~We evaluate both open and closed models and show that they struggle with compositional reasoning.
While larger models such as LLaVA-OneVision-72B (OV-72B)~\cite{onevision} tend to perform better than smaller ones (OV-7B), even the best commercial model (Gemini-1.5-Pro~\cite{gemini15flash}) achieves 49.3\% accuracy, about half that of humans at 93.0\%.
(iv)~Our experiments reveal important findings:
a)~Understanding actions is harder than agents for open models, and
b)~tests incorporating in-video negation are more challenging than text-inspired negation (\cref{subsec:eval_strictvle}).
c)~Smaller models are predisposed towards \quoteyes{} for the entailment task (\cref{subsec:eval_strictvle_posneg}).
d)~\classicvle{} hides information as entailment scores of positive and negative captions are often close to each other, likely due to subtle differences between them
(\cref{subsec:eval_classicvle}).
e)~Multiple-choice (MC) evaluation is unsuitable due to a choice bias observed even in large and closed models (\cref{subsec:eval_mc}).
f)~Finally, we show that \benchmark{} requires visual inputs and multiple frames and cannot be solved with text-only or single-frame models (\cref{subsec:blind_singleframe}).
\section{Related Work}
\label{sec:related_work}

\begin{table*}[t]
\centering
\tabcolsep=0.10cm
\footnotesize
\begin{tabular}{l c ccc c c c c}
\toprule
\multirow{2}{*}{Benchmark} & Task & \multirow{2}{*}{Comp} & In-V & Strict  & Test & Human & Video & Domain \\
& Setup & & Neg & VLE & Creation & Eval & Duration & (Source) \\
\midrule
AGQA~\cite{agqa}~\tabpub{CVPR'21} & OQA, MCQ & \cmark & \xmark & \NA{} & T, SG & \cmark & 30s & Open (ActionGenome, Charades) \\
STAR~\cite{star}~\tabpub{NeurIPSDB'21} & MCQ & \cmark & \cmark & \NA{} & H, T, SG & \xmark & 30s & Indoor (Charades) \\
ContrastSets~\cite{park2022exposing}~\tabpub{NAACL'22} & MCQ & \cmark & \xmark & \NA{} & H, T, LLM & \cmark & - & Mixed (MSR-VTT, LSMDC) \\
TestOfTime~\cite{testoftime}~\tabpub{CVPR'23} & E & \xmark & \xmark & \xmark & T & \xmark & 5-30s & Open (TEMPO, ANet Cap., Charades) \\
Perception Test~\cite{perception_test}~\tabpub{NeurIPSDB'23} & MCQ & \xmark & \xmark & \NA{} & H & \cmark & 23s & Indoor (Manual) \\
Cinepile~\cite{cinepile}~\tabpub{CVPRW'24} & MCQ & \xmark & \xmark & \NA{} & H, GPT-4 & \cmark & 2-3m & Movies (MovieClips channel) \\
VideoCon~\cite{videocon}~\tabpub{CVPR'24} & E, OQA & \cmark & \xmark & \xmark & H, PaLM-2 & \xmark & 10-30s & Open (MSR-VTT, VATEX, TEMPO) \\
SEED-Bench-2~\cite{seedbench2}~\tabpub{CVPR'24} & MCQ & \xmark & \xmark & \NA{} & H, GPT-4 & \xmark & - & Open (Charades, SSV2, EK100) \\
MV-Bench~\cite{mvbench}~\tabpub{CVPR'24} & MCQ & \cmark & \xmark & \NA{} & T, ChatGPT & \xmark & 5-35s & Mixed (Charades-STA, MoVQA, $_{+9}$) \\
TempCompass~\cite{tempcompass}~\tabpub{ACLFindings'24} & MCQ, E, VC & \cmark & \xmark & \xmark & H, GPT-3.5 & \cmark & 30s & Open (ShutterStock) \\
MMBench-Video~\cite{mmbenchvideo}~\tabpub{NeurIPSDB'24} & OQA & \xmark & \xmark & \NA{} & H & \xmark & 30s-6m & Open (YouTube) \\
VITATECS~\cite{vitatecs}~\tabpub{ECCV'24} & E & \cmark & \xmark & \xmark & H, GPT-3.5 & \cmark & 10s & Open (MSRVTT, VATEX) \\
\midrule
CVRR-ES~\cite{cvrr}~\tabpub{arXiv-2405} & OQA & \xmark & \xmark & \NA{} & H, GPT-3.5 & \cmark & 2-183s & Open (SSV2, CATER, $_{+5}$) \\ 
Video-MME~\cite{videomme}~\tabpub{arXiv-2405} & MCQ & \xmark & \cmark & \NA{} & H & \xmark & 11s-1h & Open (YouTube) \\ 
VideoVista~\cite{videovista}~\tabpub{arXiv-2406} & MCQ & \xmark & \xmark & \NA{} & T, GPT-4, GPT-4o & \xmark & 131s & Mixed (Panda-70M) \\ 
Vinoground~\cite{vinoground}~\tabpub{arXiv-2410} & E & \cmark & \xmark & \xmark & H, GPT-4 & \cmark & 10s & Open (VATEX) \\ 
TVBench~\cite{tvbench}~\tabpub{arXiv-2410} & MCQ & \cmark & \xmark & \NA{} & T & \xmark & - & Mixed (STAR, CLEVRER, $_{+6}$) \\ 
\midrule
VELOCITI (Ours) & E & \cmark & \cmark & \cmark & H, T,  LLM & \cmark & 10s & Movies (VidSitu) \\

\bottomrule
\end{tabular}
\vspace{-2mm}
\caption{ 
We review video-language benchmarks and highlight key differences to \benchmark{}.
Benchmarks use various `Task Setups':
Entailment (E),
Multiple Choice (MCQ),
Open-ended Question-Answering (OQA), and
Video Captioning (VC).
We compare \benchmark{} against benchmarks that 
test Compositionality (`Comp') or
have In-Video Negation (`In-V Neg').
In `\strictvle{}', benchmarks not adopting VLE are marked not applicable (\NA{}). 
Acronyms in the `Test Creation' column are:
template (T), scene graph (SG), open large language model (LLM), and human (H).
The `Domains' are of 3 types: Open (natural videos), Movies, and Mixed (natural \& movies).
Different from others, \benchmark{} introduces \strictvle{} and features tests with negative captions created from entities appearing in the same video. 
}
\vspace{-4mm}
\label{tab:BenchmarkComparison}
\end{table*}

Several benchmarks exist to evaluate image-language compositionality (Winoground~\cite{winoground}, COLA~\cite{cola}, 
MMVP~\cite{mmvp},
and others~\cite{vlchecklist, crepe, aro, sugarcrepe, pope}).
They require identifying the correct caption among distractors, exposing models failure to bind concepts~\cite{clipbind}.
We focus on short complex videos.

\paragraph{Video-language benchmarks}
broadly related to our work are presented in \cref{tab:BenchmarkComparison}.
We discuss differences to closely related work here.
Among previous benchmarks that study compositional reasoning, our work differs due to the
(i)~emphasis on a test suite that provides disentangled understanding of agents and actions across multiple events;
(ii)~task formulation as \textit{strict} video-language entailment 
(unlike TestOfTime~\cite{testoftime}, VideoCon~\cite{videocon}, VITATECS~\cite{vitatecs}, Vinoground~\cite{vinoground});
(iii)~explicit use of text-inspired \textit{and} in-video negation
(unlike TestOfTime~\cite{testoftime}, VideoCon~\cite{videocon}, MV-Bench~\cite{mvbench}, VITATECS~\cite{vitatecs}); and
(iv)~use of short complex videos (\eg~compared to indoor, single-agent Charades~\cite{charades} in AGQA~\cite{agqa}, STAR~\cite{star}).

While contrast captions are a popular strategy~\cite{videocon, park2022exposing, testoftime, mvbench, vitatecs, vinoground, tvbench}, the structured SRL annotations used in \benchmark{} facilitate evaluating specific aspects of a model's capabilities.
Further, different from comprehensive benchmarks that evaluate holistic video understanding (\eg~SEED-Bench-2~\cite{seedbench2}, CVRR-ES~\cite{cvrr}, Video-MME~\cite{videomme}), we focus on the fundamental ability of compositional understanding and highlight major shortcomings.
Importantly, our tests are designed to prevent text-only and single-frame models from solving them (validated empirically), guarding against issues highlighted by ATP~\cite{revisit_video} and recently TVBench~\cite{tvbench}.

\paragraph{Video-Language Entailment (VLE)} 
is posed as a binary classification task~\cite{sanyalmachines}.
Given a premise (the video) and a hypothesis (the caption), a model should determine if the hypothesis logically follows (entails) from the premise.
Entailment was first used with images in~\cite{visualentailmentimages} and adopted by~\cite{testoftime, videocon, vitatecs} for videos.
Given the rise of Vision LLMs, entailment scores are computed using the likelihood over specific words in the vocabulary~\cite{naturalbench, vqascore, videocon}.
However, most works only require that the positive caption scores higher than the negative~\cite{visualentailmentimages, vitatecs}, or with a margin~\cite{naturalbench}. 
We propose a more demanding form, \textit{\strictvle{}}, that unlike \classicvle{}, is applicable to both open and closed models (without likelihood scores).
Specifically, we independently require that the model
entails the positive caption \textit{and}
does not entail the negative caption.
While this looks simple, we find that models do not sufficiently distinguish positive and negative captions and tend to answer \quoteyes{} for both.

\section{VELOCITI Benchmark}
\label{sec:benchmark}

\begin{figure*}
\centering

\includegraphics[width=0.98\linewidth]{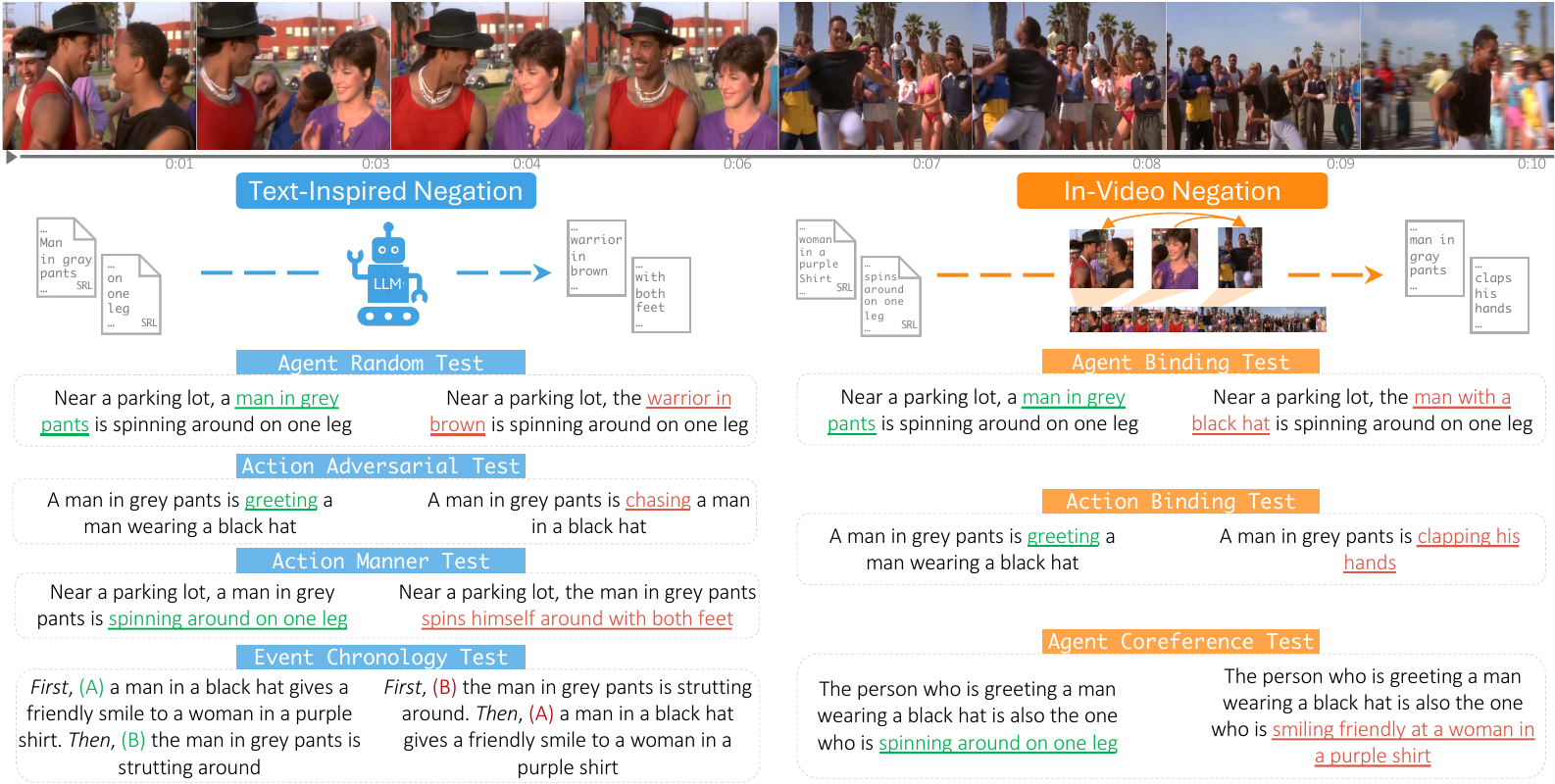}
\vspace{-1mm}
\caption{
\benchmark{} evaluates \vllm{}s' video-language entailment capabilities on complex movie clips with dense semantic role label (SRL) annotations from the VidSitu dataset~\cite{vidsitu}.
Positive and negative captions are shown side-by-side for each test with the key difference highlighted with green/red.
Negative captions are created by
(i)~manipulating text using an LLM (\bluetext{Text-Inspired Negation}) or
(ii)~replacing agents or actions by others that appear in the same video (\orangetext{In-Video Negation}).
We also demonstrate how the same positive caption can be used to create negative captions differently
(see Agent Random \vs~Agent Binding test; or
Action Adversarial \vs~Action Binding test).
Each test evaluates models for different facets of compositional reasoning as described in \cref{subsec:tests}.
The \SI{10}{\second} video clip used in this example can be viewed here: \url{https://www.youtube.com/embed/bt6-F11LZsQ?start=25&end=35}.
}
\vspace{-4mm}
\label{fig:tests}
\end{figure*}

We evaluate compositional reasoning using dynamic \SI{10}{\second} movie clips and SRL annotations from the VidSitu dataset.
We propose \textit{seven tests} to evaluate model's comprehension of agents and actions across multiple events through time.
Each test consists of $\{V, C^+, C^-\}$:
video clip $V$,
a positive caption $C^+$ that is aligned with a part of the video, and
a negative caption $C^-$ that is \textit{not} aligned to the video.
We require models to \textit{independently} assess each caption and classify them as
$V$ entails $C^+$ \textit{and} $V$ does not entail $C^-$.

\subsection{From SRL to a Video-Caption Pair}
\label{subsec:poscap}

In VidSitu, videos are divided into five \SI{2}{\second} events~\cite{vidsitu} (total \SI{10}{\second} duration).
Each event is annotated by the most salient action and the corresponding SRL capturing:
\textit{who} is doing the action (agent),
\textit{with / to whom} (patient or receiver),
\textit{with what} (instrument, if applicable),
\textit{where} (scene or location),
\textit{how} (manner or adverb), and
\textit{why} (purpose).

We use an open \llm{} (LLaMA-3~\cite{llama3}) to convert the structured SRL dictionary of each event into a caption.
The \llm{} is prompted to combine the atomic concepts into a fluent caption (prompt in~\cref{fig:poscap_prompt}).
We filter 864 videos from the validation set and generate 3101 high-quality $(V, C)$ pairs with captions that are faithful to the SRL annotations.
Depending on the test, these captions are directly used as $C^+$ or used to \emph{form} $C^+$.

Note, movie events are not bounded by \SI{2}{\second} intervals and the SRL annotations may spill into the neighboring events.
Thus, we make a conscious choice to pair the caption $C$ with the entire \SI{10}{\second} video $V$.
To correctly decide whether $V$ entails $C$, a model needs to implicitly \textit{localize} to the appropriate temporal region in the video.
This prevents a single frame bias as reported by Atemporal Probe in~\cite{revisit_video}.

\subsection{\benchmark{} Tests}
\label{subsec:tests}

We motivate and describe the seven tests below.
\cref{fig:tests} shows an example of each test grouped based on the process used to create $C^-$:
(i)~\bluetext{\textit{text-inspired negation}} typically creates $C^-$ without looking at the video; and
(ii)~\orangetext{\textit{in-video negation}},
a key contribution of our work, uses a different entity appearing in the same video to create $C^-$.
Both are important as they help us identify pitfalls of current models.

\paragraph{0. \control.}
We start with a control test to establish a baseline understanding.
Here, $C^+$ is as described in \cref{subsec:poscap} and $C^-$ is simply a positive caption of some other random video, making it easily discernible.

\paragraph{1. \argen.}
$C^-$ is created by replacing the correct \textit{agent} with another agent that does not appear in the video ($C^+$ is as above).
Solving this test requires a model to implicitly localize the event based on the action and identify who is present/absent in the video.
We ensure that the replacing agent is not a hypernym (\eg, ``man in a shirt" is not replaced by ``man'').
The SRL dictionary is updated with the random agent and the LLM generates $C^-$.

\paragraph{2. \arghn}
also replaces the agent.
Different from above, the replaced agent is chosen from the \textit{same video} making it an \orangetext{in-video negation}.
This subtle difference requires models to identify the correct agent and bind or associate it with the event description.
Models cannot rely purely on presence/absence to solve this task.
\cref{fig:tests} shows how the same $C^+$ can be modified to create both agent tests.
Similar to above, the LLM generates $C^-$.

\paragraph{3. \coref.}
Coreference groups two or more phrases that refer to the same entity~\cite{jurafsky_speech}.
In a video, an agent can be referred to by their actions, \eg in~\cref{fig:tests}, the agent: \textit{man in grey pants} is referred by: \textit{the person who is} (i)~\textit{greeting a man wearing a black hat} or
(ii)~\textit{spinning around on one leg}.
To create this test, we identify videos with the same person acting in two or more events and construct two references for that person.
$C^+$ is formed by combining the referring expressions of the \textit{same} agent,
while $C^-$ combines referring expressions of \textit{different} agents.
This test also features complex \orangetext{in-video negation} as all concepts mentioned in both $C^+$ and $C^-$ appear in the video.
The captions are created using the template: \textit{The person who is [Event A] is also the one who is [Event B].}
Solving this test requires models to associate the correct interactions of an agent across two events.
Since the agent description is masked by \textit{the person}, a model
requires multi-level compositional reasoning, making this test particularly challenging.

\paragraph{4. \verbgen.}
$C^+$ is as described in \cref{subsec:poscap} and $C^-$ is created by replacing the \textit{action} with an adversarial alternative (a plausible action determined through the text description) that does not appear in the video.
Solving this test requires
identifying the action that the agent is performing.
Given the SRL dictionary, the LLM is prompted to first generate the adversarial action followed by $C^-$.

\paragraph{5. \manner{}}
typically features a $C^+$ that includes an adverb, emotion, or facial expression.
$C^-$ is generated by replacing this manner with a contradictory yet plausible alternative.
Solving this test is challenging as it requires understanding subtle variations in an action.
Similar to the test above, the LLM is prompted to first generate the contrasting manner followed by $C^-$.

\paragraph{6. \verbret{}.}
Here, $C^-$ is created by retaining the agent from $C^+$ and swapping the action and its modifiers with those from a different event within the \textit{same video}.
Solving this test requires models to localize events where the agent appears and bind them with the correct action.
This is another test with \orangetext{in-video negation} as actions described in both $C^+$ and $C^-$ appear in the video.
To create $C^-$, we identify an event in the same video with a different action performed by a different agent.
Next, we replace the SRL dictionary of $C^+$ with the action (and relevant modifiers) and prompt the LLM to generate $C^-$.

\paragraph{7. \order{}.}
\label{sec:chronology}
Our final test studies a model's ability to confirm whether the video and caption follow the same event progression.
Multiple events descriptions can be related through time using \textit{before, after, first, then}~\cite{testoftime}.
$C^+$ is created by concatenating event captions (from \cref{subsec:poscap}) with the template \textit{``First, [Event A]. Then, [Event B]."} where event A \textit{precedes} B.
$C^-$ simply reverses them to \textit{``First, [Event B]. Then, [Event A]."}
The events are sampled at least \SI{2}{\second} apart to prevent a chance of overlap.

\paragraph{Quality control.}
All test samples in \benchmark{} are verified by humans to ensure that $C^+$ aligns with the video and $C^-$ is misaligned.
The number of samples in each test and other details are presented in~\cref{tab:quality_control}.

\section{StrictVLE Evaluation Metric}
\label{sec:strictvle_metric}

We adopt Video-Language Entailment (VLE) as the evaluation scheme for \benchmark{}.
Given an instruction $I$ containing a video $V$ and a caption $C$, model $M$ is prompted to answer whether the video entails the caption through \quoteyes{}/\quoteno{}.
We define the entailment score similar to~\cite{sanyalmachines}:
\begin{equation}
e(V,C) = \frac{p_M(\text{\quoteyes{}}|I(V,C))}{p_M(\text{\quoteyes{}}|I(V,C))+p_M(\text{\quoteno{}}|I(V,C))} ,
\end{equation} where $p_M$ denotes the model's probability distribution over the entire vocabulary.

\paragraph{\classicvle{}}~\cite{visualentailmentimages, vitatecs}
considers that a model is correct when $e(V, C^+) > e(V, C^-)$.
The random accuracy is $50\%$.

\paragraph{Narrative example.}
Consider a simple video of a red traffic light.
Let $C^{+}$, ``The
traffic light is red'' score $p(\text{\quoteyes{}}){=}0.7$,
$p(\text{\quoteno{}}){=}0.3$;
and $C^{-}$, ``The traffic light is green'' score
$p(\text{\quoteyes{}}){=}0.6$,
$p(\text{\quoteno{}}){=}0.4$.
As $e(V, C^+) > e(V, C^-)$, \classicvle{} considers this as a correct prediction.
However, with greedy decoding (constrained to `Yes' and `No'), the model will predict `Yes' for $C^-$, which is objectively incorrect, and in this example scenario, dangerous.

\paragraph{\strictvle{}.}
As models improve, it is important for our community to hold them to higher standards.
We argue that relative ordering of the entailment scores is insufficient and we propose \strictvle{} that requires models to predict \quoteyes{} for $C^+$ \textit{and} \quoteno{} for the corresponding $C^-$.
Specifically, \strictvle{} considers a sample correct iff $e(V, C^+) {>} 0.5 \land e(V, C^-) {<} 0.5$%
\footnote{Note, this is different from the Winoground~\cite{winoground} setup that has:
(i)~2 images/videos and 2 captions; and
(ii)~still uses relative scoring.}
and has a random chance accuracy of 25\%.
The threshold 0.5 arises naturally, and is equivalent to greedy decoding of \quoteyes{}/\quoteno{} in the response.
This equivalence means \strictvle{} also works on closed models without access to $p_M$.

\paragraph{Relation to multiple-choice} (MC).
While VLE performs \textit{independent} evaluation of $C^+$ and $C^-$,
MC provides \textit{both} captions to the model at once.
Seeing both captions makes the task easier as the model needs to predict the \textit{more likely} option rather than independently assess each one (similar to \classicvle{}).
In \cref{subsec:eval_mc} we reveal biases of MC evaluation and show that our proposed \strictvle{} is preferable.

\section{Results and Discussion}
\label{sec:results_analysis}

We evaluate open and closed \vllm{}s on \benchmark{}.
First, we present results with \strictvle{}, our primary evaluation strategy, and analyze entailment scores.
We also compare results with \classicvle{} and MC, discussing some evaluation pitfalls.
Finally, we evaluate blind and single-frame models to check for bias in the benchmark.

\paragraph{Models.}
We evaluate multiple open \vllm{}s:
PLLaVA~\cite{pllava},
Video-LLaVA (V-LLaVA)~\cite{videollava},
Owl-Con~\cite{videocon}, 
Qwen2-VL (QVL)~\cite{qwen2VL}, and
LLaVA-OneVision (OV)~\cite{onevision};
closed models
\geminiflash{} (Gem-1.5F), Gemini-1.5-Pro (Gem-1.5P)~\cite{gemini15flash},
\gpt{}~\cite{gpt4o};
and humans.
Due to compute and cost constraints, we evaluate QVL-72B (at native video resolution), closed models, and humans on a subset of 150 samples from each test (created once by random selection).
More details in~\cref{suppsubsec:human_evals}.

\subsection{Evaluation with StrictVLE}
\label{subsec:eval_strictvle}

We report model performance in \cref{tab:entail_strictvle} and discuss various facets of model understanding highlighted in \cref{fig:velociti_venn} (agents, actions, multiple events, and negation strategies).

\paragraph{Control \vs~\benchmark{} average.}
Old open models (P-LLaVA, Owl-Con) struggle on the \strictvle{} setup as they have a strong bias to predict \quoteyes{}.
This leads to poor performance on the control test and the benchmark average (first and last column).
V-LLaVA and subsequent models, QVL and OV, obtain decent accuracies on the control test, ranging from 65-85\%.
Compared to the control tests, performance dips strongly on the benchmark average, with the best model, OV-72B obtaining 43.2\% accuracy (36.2\% lower than control).
On the benchmark subset, while \gpt{} posts a 46.2\% accuracy on average, it performs poorly on the control tests (analysis in \cref{subsec:eval_strictvle_posneg}).
Inversely, Gem-1.5F achieves 91.9\% on the control tests, but is close to random on the benchmark (23.9\%).
The best model, Gem-1.5P, achieves 49.3\%, far from human performance at 93.0\%.
\benchmark{} is a challenging benchmark and exposes lack of reasoning in both open and closed \vllm{}s.

\paragraph{Agent understanding tests}
include the \argen{} (\argensh{}), \arghn{} (\arghnsh{}), and \coref{} (\corefsh{}).
Broadly, they evaluate a model's ability to understand the \textit{doer} of the actions in the videos.
Compared to \argensh{}, models show worse performance on \arghnsh{} and \corefsh{}.
For example, the best performing OV-72B, achieves 63.7\%, 45.4\%, and 38.6\% accuracy respectively.
\argensh{} requires verifying the \textit{presence} of the agent,
\arghnsh{} requires disambiguating between people present in the video and \textit{binding} the correct person with the event description, and
\corefsh{} needs resolving identity across \textit{multiple events}.
This proves the difficulty of in-video negation.
We also note that OV-7B performs worse than OV-72B on complex tests (\corefsh{}, 8.0\% \vs~38.6\%) indicating that multi-level reasoning is slightly better with larger LLMs.

\begin{table}[t]
\centering
\small
\tabcolsep=1.0mm
\begin{tabular}{@{}l ccccccc c c@{}}
\toprule
\multirow{2}{*}{Model} & \multirow{2}{*}{Ctrl} & Ag & Ag & Ag & Act & Act & Act & Ev & \multirow{2}{*}{Avg}  \\
& & Rand & Bind & Cref & Adv & Man & Bind & Chr \\
\midrule
Random  & 25.0 & 25.0 & 25.0 & 25.0 & 25.0 & 25.0 & 25.0 & 25.0 & 25.0 \\
\midrule
P-LLaVA & 1.3  & 0.0  & 0.0  & 0.0  & 0.0  & 0.0  & 0.0  & 0.0  & 0.0  \\
Owl-Con & 24.3 & 3.4 & 0.7  & 0.0  & 4.3  & 2.8  & 0.6  & 0.1  & 1.7  \\
V-LLaVA & 65.8 & 16.4 & 7.6  & 0.3  & 8.7  & 3.3  & 10.6 & 3.9  & 7.3 \\
\midrule
QVL-7B  & \textbf{84.6} & 39.1 & 13.5 & 6.5  & 17.8 & 17.5 & 16.4 & 0.4  & 15.9  \\
OV-7B   & 81.6 & 56.7 & 32.9 & 8.0  &  29.7 & \textbf{30.6} & 36.4 & 30.5 & 32.1  \\
OV-72B  & 79.3 & \textbf{63.7}& \textbf{45.4} & \textbf{38.6}& \textbf{33.1} & 29.3 & \textbf{45.1}& \textbf{46.5} & \textbf{43.1}  \\
\midrule
\rowcolor[gray]{.95}
\multicolumn{10}{c}{\benchmark{} Subset} \\
Gem-1.5F & \textbf{91.9} & 56.4 & 23.8 & 4.7  & 32.9 & 21.6 & 25.0 & 2.7  & 23.9  \\
QVL-72B  & 82.7 & 56.0 & 29.3 & 35.3 & 30.0 & 24.0 & 35.3 & 1.3  & 30.2  \\
OV-72B   & 81.3 & \textbf{64.0} & 46.7 & \textbf{41.3} & 30.7 & 32.7 & 46.0 & 50.0 & 44.5  \\
GPT-4o   & 63.3 & 54.7 & 44.7 & 40.7 & \textbf{55.0}& 42.0 & \textbf{54.0}& 32.2 & 46.2  \\
Gem-1.5P & 74.3 & 60.1 & \textbf{49.7} & 36.7 & 52.3 & \textbf{43.5} & 52.3 & \textbf{50.3} & \textbf{49.3} \\

\hdashline
Human    & -    & 91.5 & 92.9 & 92.6 & 92.9 & 89.9 & 91.5 & 100. & 93.0 \\
\bottomrule
\end{tabular}
\vspace{-2.5mm}
\caption{Results on \benchmark{} using the \textbf{\strictvle{}} evaluation strategy.
The tests are abbreviated as
Ctrl (Control),
\argensh{} (Agent Random), 
\arghnsh{} (Agent Binding),
\corefsh{} (Agent Coreference),
\verbgensh{} (Action Adversarial),
\mannersh{} (Action Manner),
\verbretsh{} (Action Binding),
\ordersh{} (Event Chronology).
Avg reports the average accuracy on the 7 tests of \benchmark{}.
All models show a large gap to human performance.
}
\vspace{-4.5mm}
\label{tab:entail_strictvle}
\end{table}

\paragraph{Action understanding tests}
include the \verbgen{} (\verbgensh{}), \manner{} (\mannersh{}), and \verbret{} (\verbretsh{}).
These evaluate the model's understanding of actions and/or its modifiers.
We observe that OV-72B scores worse on action tests (35.8\% average over the 3 tests) as compared to agent tests (49.2\%), while \gpt{} achieves a balanced performance of 50.3\% and 46.7\% respectively.
While \verbgensh{} is easier than \verbretsh{} for most models, OV shows inverted results.
Further, subtle variations in actions are not captured by most models and \mannersh{} is a challenging test with OV-72B at 29.3\% and Gem-1.5P posting the highest accuracy of 43.5\%.

\paragraph{Multi-event understanding tests.}
As \arghnsh{} and \verbretsh{} adopt in-video negation, they require some level of multi-event reasoning, but are ignored in this discussion.
Instead, we focus on \coref{} (\corefsh{}) and \order{} (\ordersh{}) as they have multiple events in both captions.
Time and event order are critical to video comprehension.
However, \vllm{}s are still poor at the \ordersh{} test that requires establishing the relative order of two events.
Apart from OV-72B (46.5\%) and Gem-1.5P (subset, 50.3\%), all models are comparable to or worse than random.
This is likely as all entities mentioned in both captions are present in the video.
\corefsh{} fairs slightly better, with more models showing performance better than random: QVL-72B (35.3\%), OV-72B (38.6\%), Gem-1.5P (36.7\%), GPT-4o (40.7\%).
However, it is concerning that the smaller OV-7B model collapses on these tests (\corefsh{} 8.0\%, \ordersh{} 30.5\%).
Both tests highlight challenges of reasoning across multiple events in \vllm{}s.

\paragraph{Negation strategies.}
Finally, we observe that tests adopting in-video negation and requiring associations are harder than text-inspired negation.
For \gpt{} that achieves balanced accuracy on agent and action understanding, we observe a 5.5\% drop in performance (54.9\% \argensh{}+\verbgensh{} to 49.4\% \arghnsh{}+\verbretsh{})%
\footnote{The chosen tests provide a head-to-head comparison of text-inspired \vs~in-video negation with the same positive caption as seen in \cref{fig:tests}.}.
Solving tests with in-video negation requires reasoning as it is insufficient to only check presence of entities (all entities from both captions appear in the video).
Models need to go beyond detecting the agent and action, and learn to associate them correctly.

\paragraph{Qualitative analysis.}
Example predictions of OV-72B for each test are in~\cref{fig:qa_strictvle},~\cref{fig:qa_classicvle},~\cref{fig:qa_incorrect}.

\begin{figure*}
\centering
\includegraphics[width=1\linewidth]{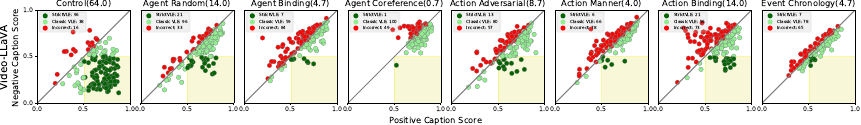}
\includegraphics[width=1\linewidth]{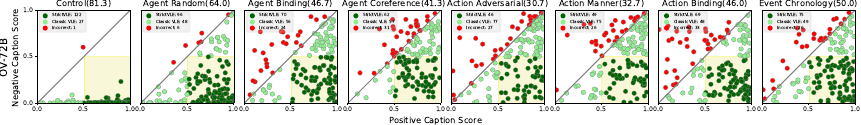}
\vspace{-5mm}
\caption{Scatter plot of entailment scores $e(V, C^+)$ (x-axis) and $e(V, C^-)$ (y-axis) for all tests in \benchmark{} subset.
We visualize the scores for Video-LLaVA (top) and OV-72B (bottom).
\classicvle{} calls a sample correct in the region below the diagonal (\textcolor{scatcorrect}{light green}).
Instead, \strictvle{} requires the dots to lie in the yellow bottom-right quadrant (\textcolor{scatstrict}{dark green}).
Finally, samples whose points are above the diagonal are wrong for both VLE metrics (\textcolor{scatincorrect}{red}).
While recent models have improved, older models concentrate near the diagonal and in the top-right \quoteyes{} quadrant for both captions.
The legend includes the actual number of points (please zoom in).
Figure is best seen in color.
}
\vspace{-3mm}
\label{fig:scatter}
\end{figure*}

\subsection{Analyzing Entailment Scores for \strictvle{}}
\label{subsec:eval_strictvle_posneg}

We analyze whether a model is better at classifying $C^+$ or $C^-$ in \cref{tab:strictVLE_analysis}.
The first number in each table cell corresponds to the accuracy of positive captions, while the second number is the accuracy of negative captions when the positive caption was correct.
We see an interesting trend.
As the model size increases, the positive caption accuracy decreases (85.9\% $\rightarrow$ 80.4\%) and negative caption accuracy increases (38.1\% $\rightarrow$ 53.6\%).
This holds for both variants: OV-7B to OV-72B and QVL-7B to QVL-72B (although on a subset).
Small models are eager to say \quoteyes{} for both captions, while larger models reason better.
A similar trend is seen on the control tests for the 7B and 72B models.
However, negative caption accuracies are far higher, confirming why control tests are easier compared to our benchmark.

\begin{table}[t]
\centering
\small
\tabcolsep=0.10cm
\begin{minipage}{0.55\linewidth}
\vspace{-0.3cm}
\begin{tabular}{l cc}
\toprule
Model & Control & Average \\
\midrule

OV-7B    & 83.0 / 98.3 & 85.9 / 38.1 \\
OV-72B   & 79.8 / \textbf{99.3} & 80.4 / \textbf{53.6} \\
QVL-7B   & \textbf{92.4} / 91.5 & \textbf{93.9} / 17.1 \\

\midrule
\rowcolor[gray]{.95}
\multicolumn{3}{c}{\benchmark{} Subset} \\

QVL-72B     & 84.0 / 98.4 & 85.2 / 36.4  \\
Gem-1.5F    & \textbf{93.9} / 97.8 & \textbf{95.8} / 25.4  \\
GPT-4o      & 63.0 / \textbf{100.} & 64.5 / \textbf{72.3}  \\
Gem-1.5P  & 75.0 / 99.1 & 74.0 / 66.2 \\
\bottomrule
\end{tabular}
\end{minipage}\hfill
\begin{minipage}{0.42\linewidth}
\caption{\strictvle{} Analysis.
We study a model's failure modes via positive and negative caption accuracy.
Each cell shows the fraction of:
(i)~correctly classified positive captions; and
(ii)~correctly classified negative captions among samples whose positive captions are correct.}
\end{minipage}
\vspace{-7mm}
\label{tab:strictVLE_analysis}
\end{table}

Somewhat unexpectedly, GPT-4o only achieves 64.5\% accuracy on positive captions.
But among them, it gets the highest negative caption accuracy of 72.3\%.
This hesitation to say \quoteyes{} hurts GPT-4o on the control tests as well and even though negative caption accuracy is perfect, it gets many positive captions wrong.
Similar analysis of each test (\cref{tab:strictvle_analysis_full}) shows that harder tests tend to have lower negative caption accuracies.

\subsection{Evaluation with \classicvle{}}
\label{subsec:eval_classicvle}

\begin{table}[t]
\centering
\small
\tabcolsep=1.0mm
\begin{tabular}{@{}l ccccccc c c@{}}
\toprule
\multirow{2}{*}{Model} & \multirow{2}{*}{Ctrl} & Ag & Ag & Ag & Act & Act & Act & Ev & \multirow{2}{*}{Avg}  \\
& {} & Rand & Bind & Cref & Adv & Man & Bind & Chr \\

\midrule
Random  & 50.0 & 50.0 & 50.0 & 50.0 & 50.0 & 50.0 & 50.0 & 50.0 & 50.0 \\
VERA    & 50.9 & 58.4 & 53.4 & 63.7 & 67.6 & 58.3 & 53.3 & 53.4 & 58.3 \\
\midrule
SigLIP  & 95.3 & 79.0 & 54.4 & 50.4 & 66.4 & 55.0 & 54.0 & 48.2 & 58.2 \\
ViFi-C  & 93.7 & 82.8 & 58.9 & 56.3 & 63.2 & 60.3 & 59.0 & 48.1 & 61.2 \\
Neg-C   & 93.4 & 83.5 & 55.3 & 50.4 & 61.6 & 61.1 & 52.4 & 50.1 & 59.2 \\
\midrule
P-LLaVA & 90.7 & 74.6 & 48.9 & 63.7 & 71.0 & 57.0 & 51.8 & 49.8 & 59.5 \\
V-LLaVA & 89.7 & 75.1 & 49.6 & 64.3 & 61.6 & 48.5 & 52.0 & 53.8 & 57.8 \\
Owl-Con & 90.8 & 73.2 & 49.9 & 48.1 & 72.4 & 61.8 & 52.7 & 42.5 & 57.2  \\
\midrule
QVL-7B  & 97.7 & 93.0 & 74.6 & 63.7 & 75.3 & 76.2 & 70.0 & 63.5 & 73.8\\
OV-7B   & 98.6 & 94.4 & 78.8 & 69.0 & 79.7 & 76.9 & 74.2 & \textbf{84.0} &  79.6\\
OV-72B  & \textbf{99.4} &\textbf{ 95.8} & \textbf{83.3} & \textbf{80.5} & \textbf{84.2} & \textbf{81.2} & \textbf{78.4} & 81.9 & \textbf{83.6}\\
\bottomrule
\end{tabular}
\vspace{-2mm}
\caption{Evaluation with \textbf{\classicvle{}}.
Random accuracy is 50\%.
Beyond \vllm{}s, we report results for a plausibility-evaluation model (VERA) and
contrastive models (SigLIP: ViT-SO400M-14-SigLIP-384~\cite{siglip}, ViFi-C: VIFICLIP-B16~\cite{vificlip}, and Neg-C: NegCLIP-B32~\cite{aro}).
The performance of contrastive models and older \vllm{}s is close to random.
However, recent models (\eg~OV) produce better relative entailment scores, even if they generate incorrect \quoteyes{}/\quoteno{} responses.
}
\vspace{-5mm}
\label{tab:entail_classicVLE}
\end{table}

While we recommend \strictvle{}, we present results on the \classicvle{} setup (\cref{tab:entail_classicVLE}) for completeness.
First, we present a language only baseline that evaluates if $C^+$ is more plausible than $C^-$.
VERA~\cite{vera} scores 58.3\% (close to random 50\%), confirming that language biases are insufficient to solve the tests.
Next, we evaluate CLIP-style models~\cite{siglip, vificlip, aro} that mean-pool video frames and observe a small improvement (ViFi-C 61.2\%).
New \vllm{}s such as QVL and OV (OV-72B: 83.6\%) show good improvement over older ones (\eg~P-LLaVA: 59.5\%).
However, this score is worse than 99.4\% on the easy control tests.
Even with a relaxed metric, OV-72B gets every sixth sample wrong.
The trends for agent and action understanding are similar: \argensh{} $>$ \arghnsh{} $>$ \corefsh{}, and
QVL and OV perform better on agent than action understanding.

To further analyze entailment scores, we present scatter plots on the benchmark subset in \cref{fig:scatter}.
While OV-72B is clearly better than Video-LLaVA, it has too many points in the top-right quadrant indicating a bias to say \quoteyes{} to both captions.
For Video-LLaVA, it is concerning that scores are close to the diagonal (\ie~both $C^+$ and $C^-$ get similar entailment scores).
In fact, these plots motivate us to propose \strictvle{} and reveal problems hidden by \classicvle{}.
We visualize such plots for all models in~\cref{fig:all_scatter}.

\subsection{MC Evaluations and Choice Bias}
\label{subsec:eval_mc}

\begin{table}[t]
\centering
\small
\tabcolsep=1.0mm
\begin{tabular}{l cccc cccc}
\toprule

\multirow{2}{*}{Model} & \multicolumn{4}{c}{Control} & \multicolumn{4}{c}{Benchmark Average} \\
& A & B & Bias & A$\land$B & A & B & Bias & A$\land$B \\
\midrule
Random      & 50.0 & 50.0 &  -     & 25.0 & 50.0 & 50.0 &  -      & 25.0 \\
QVL-7B      & 94.9 & 98.5 & (+3.6) & 94.6 & 38.9 & 88.0 & (+49.1) & 38.5\\
OV-7B       & 96.0 & 99.6 & (+3.6) & 95.9 & 28.7 & 96.5 & (+67.8) & 28.7\\
OV-72B      & 99.2 & 99.4 & (+0.2) & \textbf{99.0} & 78.0 & 88.3 & (+10.3) & \textbf{76.0} \\

\midrule
\rowcolor[gray]{.95}
\multicolumn{9}{c}{\benchmark{} Subset} \\

QVL-72B     & 100. & 100. & (+0.0) & \textbf{100.} & 73.1 & 75.7 & (\phantom{0}+2.6) & 65.3 \\
OV-72B      & 100. & 100.  & (+0.0) & \textbf{100.} & 77.1 & 87.8 & (+10.7) & \textbf{75.0}  \\
Gem-1.5F    & 100. & 99.3 & (-0.7) & 99.3 & 85.1 & 73.2 & (-11.9) & 67.7 \\
GPT-4o      & 100. & 100. & (+0.0) & \textbf{100.} & 83.9 & 74.8 & (--9.1) & 68.6 \\

\bottomrule
\end{tabular}
\vspace{-2mm}
\caption{Multi-choice (MC) evaluation results.
Along with video, we provide the model both captions as A and B and ask it to pick the better aligned one.
Column headers A (or B) refer to the accuracy when A (or B) is the positive caption.
Bias is B minus A and should be close to 0.
A$\land$B involves evaluating the model twice, once with correct caption as A and again as B.
A sample is deemed correct when it picks the correct choice in both cases.
While a model's decision should be unaffected by the order in which choices are presented, we see a considerable bias.
}
\vspace{-4mm}
\label{tab:mc}
\end{table}

In this setup, we provide the video and both captions to the \vllm{} and ask it to pick the correct description (A or B, prompts in~\cref{fig:vllm_entail_mc_prompt}, \cref{fig:gpt_entail_mc_prompt}).
In \cref{tab:mc}, we report accuracy of the model where $C^+$ is option A or option B.
We see that small 7B models have a strong choice bias and pick option B more than A (49.1\% QVL-7B or 67.8\% OV-7B).
While this reduces in larger models (10.7\% OV-72B), it is still high.
Even closed models exhibit this behavior with Gem-1.5F preferring option A over B (11.9\%) and GPT-4o preferring option A over B (9.1\%).
Interestingly, this bias becomes a major issue when the tests are challenging and is negligible in the control tests that are easier.

While one could report accuracy by running the model twice, once with option A as $C^+$ and again with B as $C^+$ (referred as A$\land$B), this is tedious and the number of evaluations increases as a factorial of the number of choices.
If we compare \strictvle{} with the MC evaluation's A$\land$B score (both apply $\land$ on binary decisions and have random chance at 25\%), we observe that MC is much easier (OV-72B: 76.0\%) than StrictVLE (43.1\%, \cref{tab:entail_strictvle}).
This may be attributed to the MC setup, where a model processes both captions at once and only needs to pick the more likely option; in contrast with the \strictvle{} setup that requires independent evaluation of each caption.
Even though MC is a popular evaluation setup for many benchmarks (see~\cref{tab:BenchmarkComparison}),
the choice bias of \vllm{}s makes results difficult to interpret.
For all these reasons, \strictvle{} is preferred.

\subsection{Validating Benchmark Properties}
\label{subsec:blind_singleframe}

\begin{table}[t]
\centering
\small
\tabcolsep=1.0mm
\begin{tabular}{@{}l ccccccc c c@{}}
\toprule
\multirow{2}{*}{Model} & \multirow{2}{*}{Ctrl} & Ag & Ag & Ag & Act & Act & Act & Ev & \multirow{2}{*}{Avg} \\
& {} & Rand & Bind & Cref & Adv & Man & Bind & Chr \\

\midrule
\rowcolor[gray]{.95}
\multicolumn{10}{c}{A. Comparing against Blind Models} \\
OV \faEye     & 79.3 & 63.7 & 45.4 & 38.6 & 33.1 & 29.3 & 45.1 & 46.5 & \textbf{43.1}  \\
Q LLM & 2.2 & 2.5  & 2.2    & 5.3    & 4.1   & 1.3  & 2.7   & 7.9     & 3.7      \\
OV Blind & 6.0 & 9.3  & 6.7     & 12.7   & 10.3 & 3.3 & 6.9    & 7.4   & 8.1    \\

\midrule
\rowcolor[gray]{.95}
\multicolumn{10}{c}{B. Impact of Single Frame Input or Model} \\
OV-7B & 81.6 & 56.7 & 32.9 & 8.0  &  29.7 & 30.6 & 36.4 & 30.5 & \textbf{32.1}  \\
1 Frame & 39.6 & 31.6 & 22.3 & 18.1 & 15.5 & 13.4 & 22.1  & 10.3 & 19.0  \\
OV-7B-SI & 78.9 & 35.7 & 15.6 & 2.9  & 27.6 & 21.4 & 22.0 & 8.8  & 19.1  \\

\bottomrule
\end{tabular}
\vspace{-2mm}
\caption{\benchmark{} benchmark validation.
\textbf{Part A.}
We confirm that the model requires visual inputs.
The base LLM Qwen2-72B (Q LLM) or OV-72B without providing the video (OV Blind) perform poorly compared to OV-72B provided with video frames (OV \faEye).
\textbf{Part B.}
We also confirm that providing multiple video frames is necessary.
When OV-7B is provided a single frame chosen randomly (1 Frame) or when the video is fed to a model trained only on single images (LLaVA-OneVision-SingleImage, OV-7B-SI), the performance dips compared to showing the video at 1fps (our default strategy, OV-7B).
}
\vspace{-4mm}
\label{tab:validate_velociti}
\end{table}

We highlight some additional properties of our benchmark.

\paragraph{Evaluating blind models.}
\cref{tab:validate_velociti}A compares Qwen2-LLM (Q LLM) and OV-72B without the video inputs (OV Blind) against the default OV-72B model (here, OV \faEye{}).
We see a dramatic drop (43.1\% to 3.7\% Q LLM and 8.1\% OV Blind).
Solving tests in \benchmark{} requires visual understanding.

\paragraph{Evaluating with a single-frame.}
\cref{tab:validate_velociti}B reports results on OV-7B models.
The first row is the default setup (\cref{tab:entail_strictvle}) and is compared against:
(i)~OV-7B with a single frame input (1 Frame) chosen at random from the sampled 1fps frames.
(ii)~The OneVision team~\cite{onevision} first train an image-only model and extend it to multiple images and videos.
We evaluate their single image checkpoint while providing video inputs (OV-7B-SI).
The performance drops from 32.1\% to about 19.0\% in both cases.
This confirms that \benchmark{} requires video inputs and video models.

CoT and FPS ablations are in Appendix~\ref{suppsubsec:cot_results} and~\ref{suppsubsec:fps}.

\section{Conclusion}
\label{sec:conclusion}
We introduced \benchmark{}, a benchmark to evaluate the compositional capabilities of \vllm{}s by disentangling and assessing the comprehension of agents, actions, and their associations across multiple events.
We improved over the classic Video-Language Entailment setup that relies on relative scoring by proposing \strictvle{} that requires models to answer \quoteyes{} for the positive caption \textit{and} \quoteno{} for the negative caption.
All evaluated models, open and closed, performed poorly with a large gap to human performance.
Our experiments showed that action understanding is harder than agent understanding, and solving tests with in-video negation is harder than text-inspired ones.
We also analyzed limitations of \classicvle{} and the choice bias in multiple-choice evaluations.
Overall, our work established that compositional reasoning on short videos is still unsolved and remains challenging for \vllm{}s.

{\small
\paragraph{Acknowledgment.}
MT and ViG thank Adobe Research for supporting this project and travel.
MT thanks SERB SRG/2023/002544 for compute server and Google Cloud credits.
We thank volunteers for human evaluations.
}

\clearpage
{
\small
\bibliographystyle{ieeenat_fullname}
\bibliography{bib/longstrings, bib/main}
}

\clearpage
\maketitlesupplementary
\appendix


In this supplementary material, we discuss
\begin{enumerate}
\item Additional results and analysis, both quantitative and qualitative (\cref{suppsec:results});
\item Benchmark creation, quality control process, and some statistics (\cref{suppsec:benchmark});
\item Model prompts used in both setups: entailment and multiple-choice (\cref{suppsec:model}); and
\item Limitations (\cref{suppsec:limitations}).
\end{enumerate}

\section{Additional Results}
\label{suppsec:results}

In \cref{suppsubsec:strictvle_score_plots},
we present scatter plots of entailment scores for all models across all tests, expanding Fig. 3 from the main paper.
Next, we present the positive and negative entailment scores that are used in StrictVLE (expanding the analysis in Sec. 5.2) in \cref{suppsubsec:strictvle_analysis}.
We experiment with Chain-of-Thought prompting in \cref{suppsubsec:cot_results} and present ablations for number of sampled frames in \cref{suppsubsec:fps}.
The multiple-choice (MC) evaluation results are discussed in \cref{suppsubsec:mc_results} and
the human evaluation setup in \cref{suppsubsec:human_evals}.
Finally, we share some qualitative results of \ovseventy{} on our benchmark in \cref{subsec:qual_analysis}.

\subsection{Scatter plot of entailment scores}
\label{suppsubsec:strictvle_score_plots}

\begin{figure*}[t]
\centering
\includegraphics[width=\textwidth]
{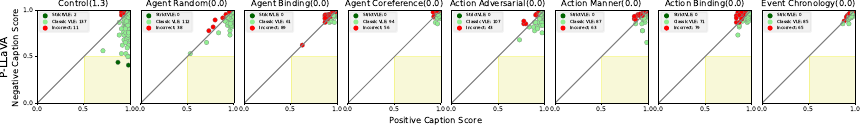}
\includegraphics[width=\textwidth]
{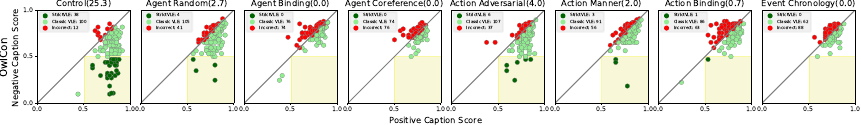}
\includegraphics[width=\textwidth]
{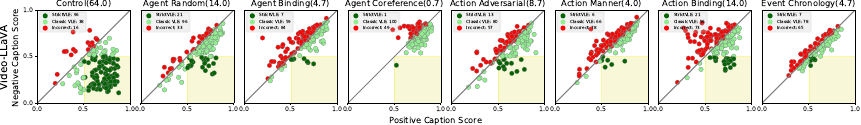}
\includegraphics[width=\textwidth]
{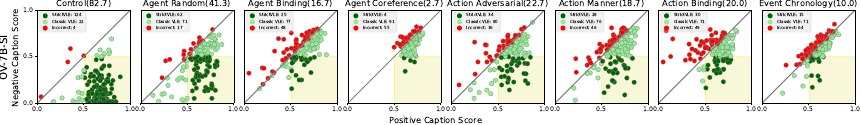}
\includegraphics[width=\textwidth]
{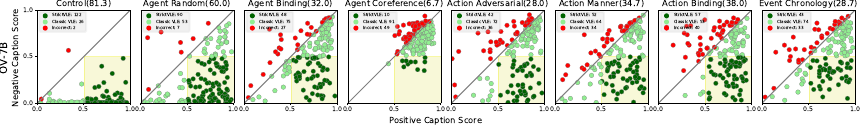}
\includegraphics[width=\textwidth]
{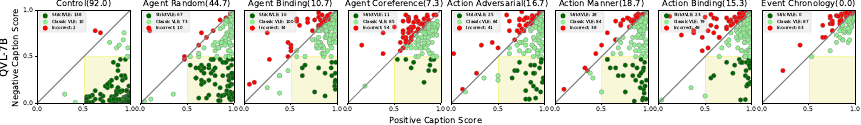}
\includegraphics[width=\textwidth]
{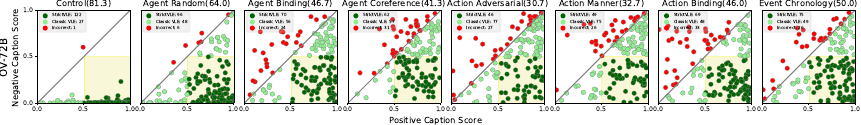}
\caption{Scatter plot of entailment scores $e(V, C^+)$ (x-axis) and $e(V, C^-)$ (y-axis) for all tests in \benchmark{}.
We visualize the scores for several models indicated in the left margin.
From top to bottom: P-LLaVA, OwlCon, Video-LLaVA, OV-7B-SI, OV-7B, QVL-7B, and OV-72B.
\classicvle{} calls a sample correct in the region below the diagonal (\textcolor{scatcorrect}{light green}).
Instead, \strictvle{} requires the dots to lie in the yellow bottom-right quadrant (\textcolor{scatstrict}{dark green}).
Finally, samples whose points are above the diagonal are wrong for both VLE metrics (\textcolor{scatincorrect}{red}).
The legend includes the actual number of points (please zoom in).
This figure is best seen in color.
}
\vspace{-4mm}
\label{fig:all_scatter}
\end{figure*}

To analyze entailment scores, we present scatter plots for all models on the benchmark subset (150 samples) in \cref{fig:all_scatter}.
The ideal scenario is when all samples lie in the bottom-right quadrant (points in {\color{scatstrict}dark green}, quadrant in light yellow), which indicates that the model confidently entails the correct caption while rejecting the negative caption, leading to a 100\% \strictvle{} accuracy.
However, in practice, we observe two undesirable cases:
(i)~the points are concentrated in the top-right quadrant, indicating a strong bias towards responding \quoteyes{} regardless of whether the caption is aligned or misaligned; and
(ii)~the points are clustered around the diagonal, indicating that the model exhibits similar confidence levels when saying \quoteyes{} to both the positive and negative captions.
Major takeaways are highlighted below:

\begin{itemize}
\item \pllava{} has most of its points concentrated in the top-right quadrant, indicating a strong bias towards responding \quoteyes{} regardless of whether the caption is positive or negative, which also explains its near 0\% \strictvle{} accuracy. 

\item \videocon{} and \vidllava{} are strongly clumped near the diagonal in the top-right quadrant (except for the \control): indicating that they tend to respond \quoteyes{} and have similar entailment scores for both the positive and negative captions.
\videocon{} appears to be worse than\vidllava{} with more points in the top-right quadrant.

\item Between \ovseven{} (OV-7B) and \ovsevensi{} (OV-7B-SI), we see that the points in OV-7B-SI are more clustered near the diagonal while \ovseven{} is more diffused except for \corefsh{}.
This is expected as it is hard for a model trained on single images to distinguish between the positive and the negative caption and nearly impossible for the \ordersh{} and \corefsh{}. 
In contrast, both models perform well on the \control{} since the replacements come from a totally different or random video, making it easier for the models to classify with sufficient confidence.

\item For \qwenvl-7B (QVL-7B) except for \control{}, the points for all the other tests are concentrated in the top-right corner while additionally being clustered near the diagonal for the \ordersh{}.
QVL-7B performs worse than OV-7B even though both models are trained using the same base Qwen2 7B parameter LLM.

\item Finally, on \ovseventy{}, we see that many points are below the diagonal and would score correct on ClassicVLE.
However, roughly half of them (on average) are in the bottom right quadrant indicating difficulty of the best model to predict \quoteyes{} for the positive caption and \quoteno{} for the negative caption respectively.

\end{itemize}

\subsection{Analyzing Entailment Scores for \strictvle{}}
\label{suppsubsec:strictvle_analysis}
Continuing from findings of Sec. 5.2 in the main paper, we analyze whether a model finds it easier to classify $C^+$ or $C^-$ in \cref{tab:strictvle_analysis_full} for all tests. 
Each cell in the table reports two numbers: the first is the accuracy of positive captions, and the second is the accuracy of negative captions when the positive caption is correct.

An interesting observation (as also noted in the main paper) is that as model size increases, the positive caption accuracy decreases while the negative caption accuracy improves.
This holds for both variants: OV-7B to OV-72B and QVL-7B to QVL-72B, and indicates that small models are eager to say \quoteyes{} for both captions, while larger models reason better.

Although \qwenvl{} (QVL) models achieve higher accuracy for positive captions than \ov{} (OV) models, the negative caption accuracy is better for OV models.
This indicates that QVL models are biased to say \quoteyes{} regardless of the captions, whereas OV models reason better and are less inclined to respond \quoteyes{}.
For QVL, specifically for the \ordersh{} test, the positive caption accuracy is very high, but the negative caption accuracy is extremely low, indicating that the QVL models are very poor at the temporal order reasoning.

While \gpt{} achieves comparatively lower accuracy on positive captions across all tests, it consistently achieves the highest accuracy for negative captions, except in the \ordersh{} test, where OV-72B performs best.
Another surprising observation is that \geminiflash{} (Gem-1.5F), despite achieving the best accuracy for positive captions, performs worse than all other models for negative captions.
This suggests that \geminiflash{} may also be responding with \quoteyes{} too often.
Additionally, both \geminiflash{} and \qwenvlseventy{} exhibit very low accuracy for negative captions in the \corefsh{} and \ordersh{} tests.

Finally, in Sec. 5.1 of the main paper, we highlight that \argensh{} $>$ \arghnsh{} $>$ \corefsh{} -- this trend is clearly observed in the negative caption accuracies presented in the Table,
and explains the poor performance of some models on \coref{} (single-digit accuracies on negative captions).

\begin{table*}[t]
\tabcolsep=1.2mm
\centering
\begin{tabular}{l c ccccccc c} 
\toprule
Model & \multirow{2}{*}{Ctrl} & Ag & Ag & Ag & Act & Act & Act & Ev & \multirow{2}{*}{Avg} \\
& & Rand & Bind & Cref & Adv & Man & Bind & Chrono \\
\midrule
OV-7B & 83.0 / 98.3 & 84.2 / 67.3 & 82.0 / 40.1 & 97.1 / \phantom{0}8.2 & 86.3 / 34.4 & 80.6 / 37.9 & 81.9 / 44.5 & 89.0 / 34.2 & 85.9 / 38.1 \\
OV-72B & 79.8 / 99.3 & 81.6 / 78.1 & 79.5 / 57.1 & 87.0 / 44.4 & 80.6 / 41.1 & 77.9 / 37.5 & 78.2 / 57.6 & 78.3 / 59.4 & 80.4 / 53.6 \\
QVL-7B & 92.4 / 91.5 & 92.8 / 42.1 & 90.7 / 14.9 & 97.6 / \phantom{0}6.6 & 94.3 / 18.9 & 93.0 / 18.8 & 91.0 / 18.1 & 98.0 / \phantom{0}0.4 & 93.9 / 17.1 \\
\midrule
\rowcolor[gray]{.95}
\multicolumn{10}{c}{\benchmark{} Subset} \\
QVL-72B & 84.0 / 98.4 & 85.3 / 65.6 & 84.0 / 34.9 & 77.3 / 45.7 & 84.0 / 35.7 & 86.7 / 27.7 & 80.7 / 43.8 & 98.7 / \phantom{0}1.4 & 85.2 / 36.4 \\
OV-72B & 81.3 / 100. & 79.3 / 80.7 & 80.7 / 57.9 & 86.7 / 47.7 & 79.3 / 38.7 & 82.0 / 39.8 & 74.7 / 61.6 & 80.7 / 62.0 & 80.5 / 55.5 \\
Gem-1.5F & 93.9 / 97.8 & 93.3 / 60.4 & 93.9 / 25.4 & 100. / \phantom{0}4.7 & 93.3 / 35.3 & 95.3 / 22.7 & 95.3 / 26.2 & 99.3 / 2.8 & 95.8 / 25.4 \\
Gem-1.5P & 75.0 / 99.1 & 72.3 / 83.2 & 75.5 / 65.8 & 71.3 / 51.4 & 72.5 / 72.2 & 79.6 / 54.7 & 75.5 / 65.8 & 71.4 / 70.5 & 74.0 / 66.2 \\
GPT-4o & 63.3 / 100.0 & 58.0 / 94.3 & 64.0 / 69.8 & 62.0 / 65.6 & 65.8 / 83.7 & 65.3 / 64.3 & 64.7 / 83.5 & 71.8 / 44.9 & 64.5 / 72.3 \\
\bottomrule
\end{tabular}
\caption{\strictvle{} Analysis for various models on all tests in \benchmark{}. Each cell of the table has two numbers.
The first is the fraction of correctly classified positive captions.
The second is the fraction of correctly classified negative captions,
among samples whose positive caption is classified correctly.
Refer to \cref{suppsubsec:strictvle_analysis} for a description.
}
\label{tab:strictvle_analysis_full}
\end{table*}

\subsection{Impact of Chain-of-Thought prompting}
\label{suppsubsec:cot_results}
We experimented with Chain-of-Thought (CoT) prompting for Gemini-1.5-Pro and OV-72B (prompt in~\cref{fig:cot_prompt}). As shown in \cref{tab:cot}, the performance reduced in both cases indicating that models are unable to reason in a step-by-step manner for such statements.
\begin{table}[t]
\centering
\tabcolsep=1.0mm
\begin{tabular}{@{}l cccccc c c@{}}
\toprule
\multirow{2}{*}{Model} & Ag & Ag & Ag & Act & Act & Act & Ev & \multirow{2}{*}{Avg}  \\
& Rand & Bind & Cref & Adv & Man & Bind & Chr \\
\midrule
\rowcolor[gray]{.95}
\multicolumn{9}{c}{OV-72B} \\
w/o CoT & 64.0 & 46.7 & 41.3 & 30.7 & 32.7 & 46.0 & 50.0 & 44.5\\
CoT & 40.0 & 28.0 & 19.3 & 26.7 & 28.0 & 32.0 & 30.0 & 29.1 \\
\rowcolor[gray]{.95}
\multicolumn{9}{c}{Gemini-1.5 Pro} \\
w/o CoT & 60.1 & 49.7 & 36.7 & 52.3 & 43.5 & 52.3 & 50.3 & 49.3 \\
CoT & 46.6 & 32.8 & 45.5 & 46.2 & 46.8 & 46.4 & 29.2 & 41.9 \\
\bottomrule
\end{tabular}
\vspace{-3mm}
\caption{ Average score on \benchmark{} subset: without and with CoT}
\vspace{-4mm}
\label{tab:cot}
\end{table}

\begin{figure}[t]
\centering
\noindent\begin{minipage}{\textwidth}
\mdfsetup{%
middlelinewidth=1pt,
backgroundcolor=cyan!10,
innerleftmargin=0.5cm,
innerrightmargin=0.5cm,
font=\small,
roundcorner=15pt}
\begin{mdframed}
\vspace{0.2em}

\textbf{System Prompt}\\

You are an AI assistant specializing in analyzing movie clips to verify captions using a Chain of Thought (CoT) approach.
Given a movie clip and a corresponding caption, your task is to determine whether the caption accurately describes the events in the clip.\\

A caption is considered accurate (``Yes") if *all applicable* of the following criteria are met:
\begin{enumerate}
\item {**Actor/Doer**: The person or entity performing the action is correctly identified.}
\item {**Attributes**: The characteristics of the actors/doers and the action itself are accurately described (\eg clothing color, size, speed).}
\item {**Instruments/Objects**: Any tools, objects, or instruments used in the action are correctly identified.}
\item {**Receiver/Patient**: The target or recipient of the action is correctly identified.}
\item {**Relationships**: The relationships between the entities involved (e.g., ``standing next to",``holding") are accurately depicted.}
\item {**Manner**: The way in which the action is performed (e.g., ``quickly", ``slowly", ``angrily") is accurately described.}
\item {**Location**: The setting or location of the scene is correctly identified.}
\item {**Clarity**: There is sufficient visual information in the clip to confidently assess the correctness of the caption.}
\item {**Event Order**: If the caption suggests a specific order of events, then the video should have events happening in the suggested order.}
\end{enumerate}
If any of the above criteria cannot be verified due to a lack of visuals, the caption should not be considered accurate.

Note that the caption is designed to represent a part of the video clip and may not explain all the events in the clip.
\vspace{1em}

Follow these steps:\\
1.  **Analysis:** Carefully examine the provided movie clip.\\
2.  **Reasoning:** Analyze the caption in relation to the clip. Break down the caption into smaller parts and determine if each part meets the accuracy criteria listed above. Detail your reasoning process within `$<$thinking$>$` tags.\\
3.  **Evaluation:** Based on your reasoning, evaluate the overall accuracy of the caption. If there is insufficient information in the clip to definitively confirm or deny the caption based on one or more criteria, explain what information is missing within `$<$reflection$>$` tags.\\
4.  **Conclusion:** Provide a clear ``Yes" or ``No" answer within `$<$output$>$` tags.\\

Use the following format:\\
$<$thinking$>$\\
$[$Detailed step-by-step reasoning, referencing the accuracy criteria. This is your internal thought process.$]$\\
$<$/thinking$>$
\vspace{0.5em}

$<$reflection$>$\\
$[$Reflections on your reasoning, including any uncertainties or missing information and which criteria could not be verified. If the caption cannot be definitively verified, explain why.$]$ \\
$<$/reflection$>$
\vspace{0.5em} \\

$<$output$>$\\
$[$ Yes or No $]$ \\
$<$/output$>$\\
\vspace{1em}
Evaluate the following caption for the accompanying movie clip:
\{caption\}

\end{mdframed}
\end{minipage}
\caption{CoT evaluation prompt.}
\label{fig:cot_prompt}
\vspace{-1em}
\end{figure}

\subsection{Impact of Increased Frame Rate}
\label{suppsubsec:fps}
We explore increasing the video sampling rate to observe if more visual information aids the model to solve the tasks in \benchmark{} to a greater extent.
For this, we sample frames at 8 fps, amounting to 80 frames for a \SI{10}{\second} video.
From \cref{tab:fps} we observe that the smaller model (OV-7B) benefits with more frames resulting in improvement across most tests, an average of +3.5\%.
Interestingly, its larger counterpart (OV-72B) performs worse with significant drops on action tests, ActAdv and ActMan, (9-11\%).
This may be due to the large context size that the model is not trained for.
Both models perform better on EvChr task.

\begin{table}[t]
\centering
\tabcolsep=1.0mm
\begin{tabular}{@{}l cccccc c c@{}}
\toprule
\multirow{2}{*}{Model} & Ag & Ag & Ag & Act & Act & Act & Ev & \multirow{2}{*}{Avg}  \\
& Rand & Bind & Cref & Adv & Man & Bind & Chr \\
\midrule
\rowcolor[gray]{.95}
\multicolumn{9}{c}{OV-7B} \\
1fps   & 56.7 & 32.9 & 8.0  &  29.7 & 30.6 & 36.4 & 30.5 & 32.1  \\
8fps & 59.3 & 34.7 & 6.0 & 34.7 & 38.3 & 33.3 & 42.0 & 35.6 \\
\rowcolor[gray]{.95}
\multicolumn{9}{c}{OV-72B} \\
1fps & 64.7 & 46.0	& 36.7 & 42.0 & 40.7 & 46.0 & 46.0 & 46.0 \\
8fps & 63.7& 45.4 & 38.6 & 33.1 & 29.3 & 45.1& 46.5 & 43.1  \\
\bottomrule
\end{tabular}
\vspace{-3mm}
\caption{Higher frame rate sampling results.}
\label{tab:fps}
\end{table}

\subsection{Multiple-Choice (MC) Evaluation: Results on each test}
\label{suppsubsec:mc_results}

\begin{table}[t]
\centering
\tabcolsep=1.5mm
\begin{tabular}{l rrrrrrrrrrrrrr}
\toprule
\multirow{2}{*}{Model} & \multicolumn{2}{c}{\argensh} & \multicolumn{2}{c}{\arghnsh} & \multicolumn{2}{c}{\corefsh} & \multicolumn{2}{c}{\verbgensh} & \multicolumn{2}{c}{\mannersh} & \multicolumn{2}{c}{\verbretsh} & \multicolumn{2}{c}{\ordersh} \\
& Bias & A$\land$B & Bias & A$\land$B & Bias & A$\land$B & Bias & A$\land$B & Bias & A$\land$B & Bias & A$\land$B & Bias & A$\land$B \\
\midrule
QVL-7B & 24.2 & 74.1 & 42.2 & 40.8 & 37.5 & 33.6 & 49.6 & 42.9 & 51.5 & 41.5 & 40.0 & 36.1 & 98.5 & 0.7 \\
OV-7B & 41.6 & 58.1 & 81.6 & 17.1 & 59.9 & 26.0 & 71.3 & 27.6 & 68.0 & 30.1 & 70.0 & 24.2 & 81.9 & 15.9 \\
OV-72B & 3.1 & 94.8 & 10.9 & 72.9 & 8.0 & 69.0 & 8.9 & 79.7 & 11.1 & 77.5 & 14.8 & 62.5 & 15.1 & 75.9 \\
\rowcolor[gray]{.95}
\multicolumn{15}{c}{\benchmark{} Subset} \\
QVL-72B & 6.0 & 88.7 & 3.3 & 64.7 & 2.7 & 60.0 & 6.7 & 68.0 & 2.0 & 74.0 & 8.6 & 54.7 & -11.3 & 47.3 \\
OV-72B & 2.7 & 95.3 & 11.4 & 75.3 & 11.3 & 67.3 & 9.3 & 76.7 & 7.3 & 77.3 & 16.0 & 61.3 & 16.7 & 72.0 \\
Gem-1.5F & -2.8 & 94.4 & -12.6 & 73.4 & 8.0 & 61.3 & -12.9 & 72.8 & -14.3 & 66.0 & 0.7 & 64.8 & -49.6 & 41.4 \\
GPT-4o & 4.1 & 92.5 & 4.7 & 79.3 & 3.4 & 60.8 & -10.1 & 77.0 & -7.0 & 74.1 & 2.0 & 70.7 & -60.8 & 25.7 \\
\bottomrule
\end{tabular}
\caption{MC evaluation results on all tests.
Along with the video, we provide the model with both captions A and B and ask it to pick the better-aligned one.  Bias is the accuracy difference between B and A options and should be close to 0.
A$\land$B involves evaluating the model twice, once with the correct caption as A and again as B. A sample is deemed correct when it picks the correct choice in both cases. While a model's decision should be unaffected by the order in which choices are presented, a considerable bias is observed.}
\label{tab:mc_results_full}
\end{table}

In the MC setup, we provide the video along with both captions to the \vllm{} and ask it to pick the correct one (A or B).
Results on the control and average over the benchmark were discussed in Sec. 5.4 of the main paper.

Now, we report results across all the tests in~\cref{tab:mc_results_full}.
For both OV and QVL models, we see that the smaller variants have a higher choice bias and tend to prefer option B. While this bias reduces in the larger variants, it is still high. Also, as expected, the accuracy of A$\land$B improves for larger variants.
We observe that harder tests (\eg~\arghnsh{} \vs~\argensh{}) tend to have a higher bias.
Among all the tests, the \ordersh{} test has the highest bias and the lowest accuracy across all the models.

Both \geminiflash{} and \gpt{} show considerable bias.
Interestingly, \gpt{} seems to prefer option A, while Gem-1.5F prefers option B.

\subsection{Human Evaluation}
\label{suppsubsec:human_evals}

\begin{figure*}
\centering
\includegraphics[width=0.95\textwidth]{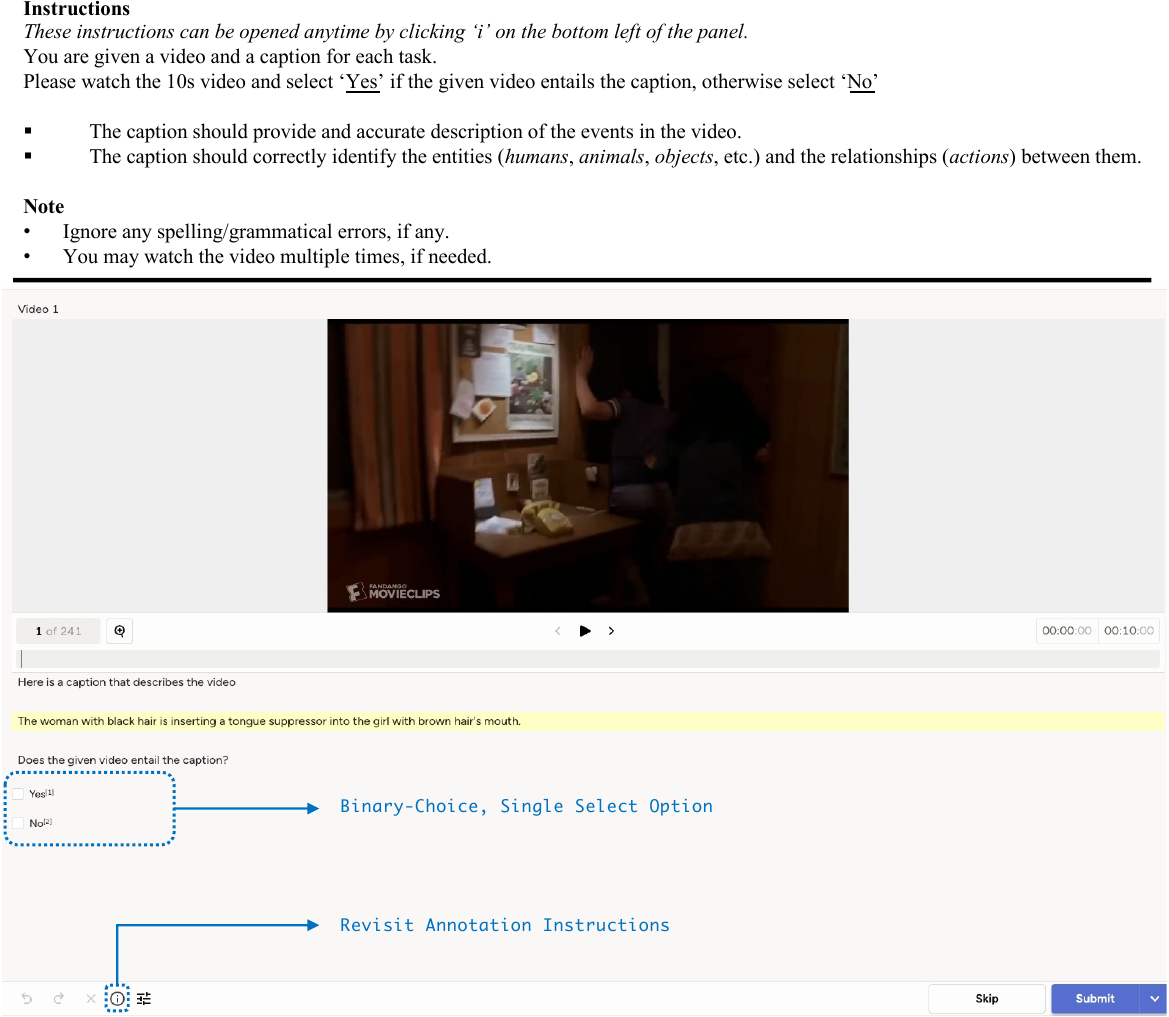}
\caption{Human Evaluation Dashboard. Instructions and interface for human evaluation for the entailment task.}
\label{fig:human_eval}
\end{figure*}

Human evaluations were conducted in a standardized manner to establish human performance in the various tasks presented in \benchmark{}. 
The evaluations included 3 volunteers who were assigned the subset (150 samples for each of the 7 tests).
This amounts to a total of 2,100 video-caption pairs (7 tests $\times$ 150 samples $\times$ 2 captions).
We use the Label Studio~\cite{labelstudio} annotation platform for this task.
To ensure fair evaluations, humans are first shown a set of instructions to ensure consistency across participants.
Next, we randomize and present non-overlapping video-caption pairs.
An example of the annotation dashboard is shown in \cref{fig:human_eval}.

\subsection{Qualitative Analysis}
\label{subsec:qual_analysis}

\begin{figure*}[t]
\centering
\includegraphics[width=0.9\linewidth]{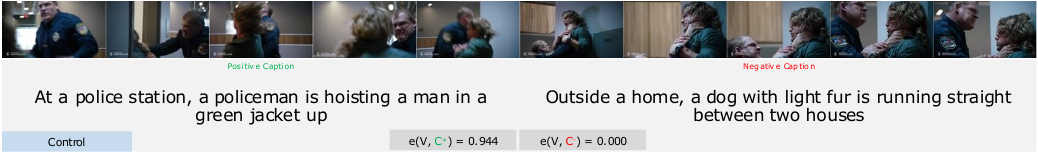}\vspace{0.3em}
\includegraphics[width=0.9\linewidth]{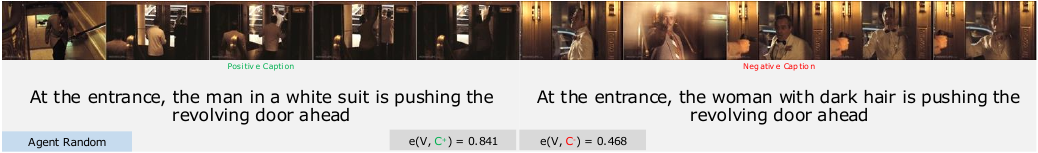}\vspace{0.3em}
\includegraphics[width=0.9\linewidth]{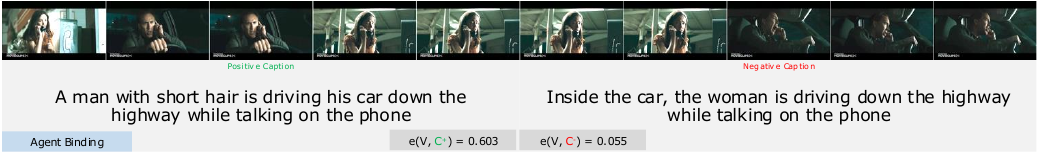}\vspace{0.3em}
\includegraphics[width=0.9\linewidth]{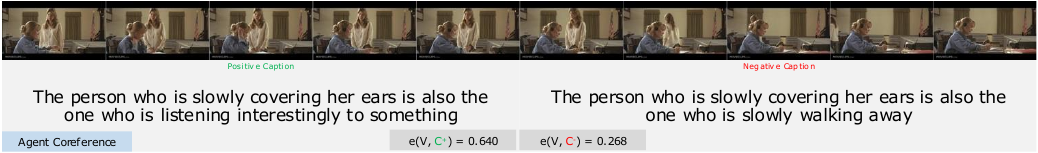}\vspace{0.3em}
\includegraphics[width=0.9\linewidth]{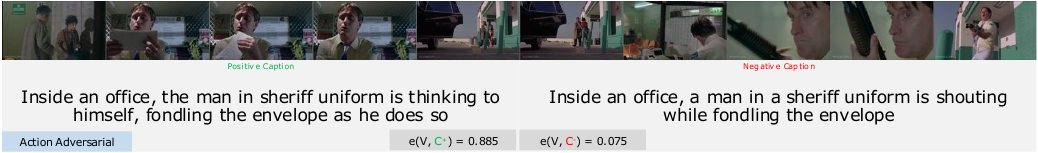}\vspace{0.3em}
\includegraphics[width=0.9\linewidth]{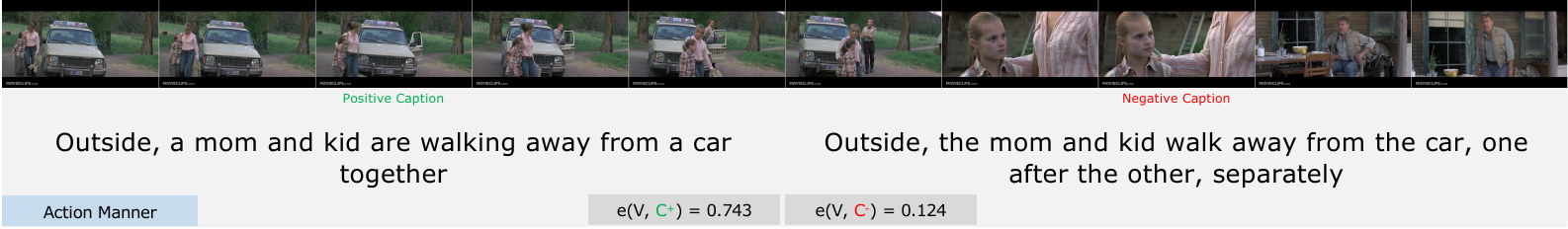}\vspace{0.3em}
\includegraphics[width=0.9\linewidth]{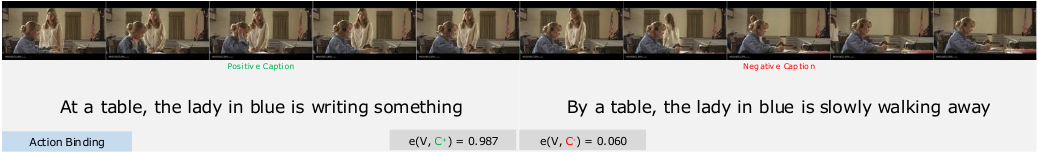}\vspace{0.3em}
\includegraphics[width=0.9\linewidth]{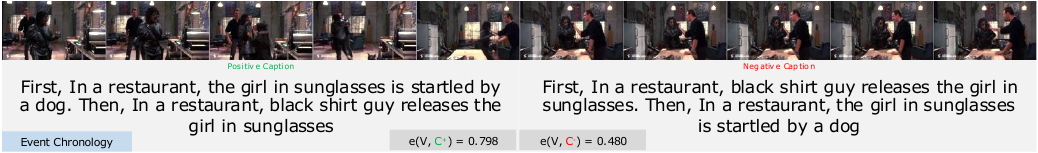}
\caption{\benchmark{} samples where OV-72B classifies the sample correctly based on the \strictvle{} criteria.
In the \qabluetext{Agent Binding} example, the scene visualizes a man and a woman talking on the phone while the man drives, $C^{-}$ changes the entity of the driver. The model is confidently able to identify that it is the man who is driving and not the woman, as the positive caption scores (0.603) much above the negative caption (0.055) while satisfying the \strictvle{} criteria. Similarly, in \qabluetext{Agent Coreference}, the scene describes two women - a woman in blue who's sitting and puts on her headphones as she begins to write, while the woman in white looks at her and eventually walks away. The $C^{-}$ interchanges the roles of these two women, and the model correctly scores the positive caption (0.640) higher than the negative caption (0.268).
}
\label{fig:qa_strictvle}
\end{figure*}

\begin{figure*}[t]
\centering
\includegraphics[width=0.9\linewidth]{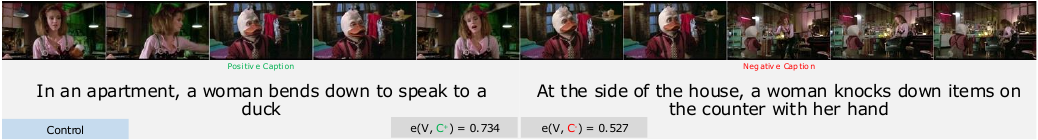}\vspace{0.3em}
\includegraphics[width=0.9\linewidth]{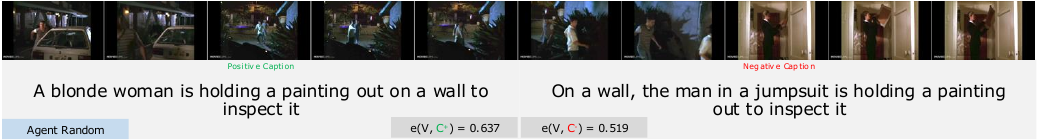}\vspace{0.3em}
\includegraphics[width=0.9\linewidth]{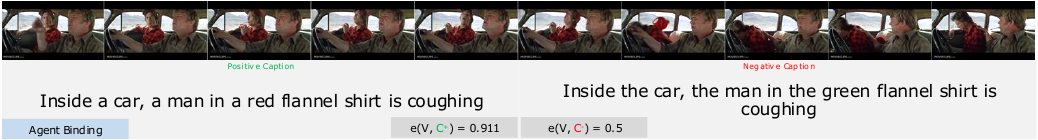}\vspace{0.3em}
\includegraphics[width=0.9\linewidth]{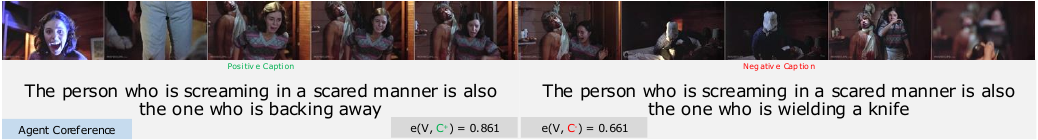}\vspace{0.3em}
\includegraphics[width=0.9\linewidth]{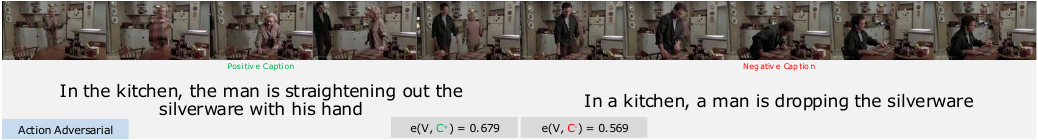}\vspace{0.3em}
\includegraphics[width=0.9\linewidth]{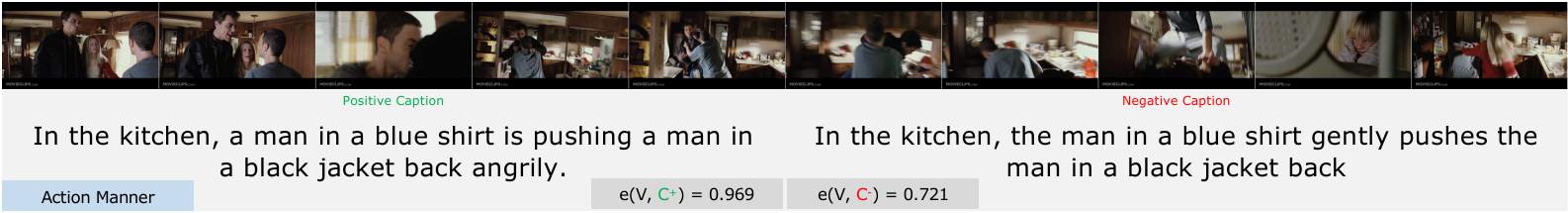}\vspace{0.3em}
\includegraphics[width=0.9\linewidth]{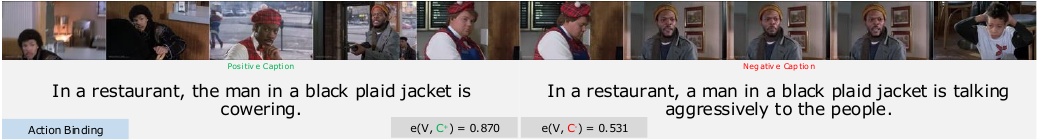}\vspace{0.3em}
\includegraphics[width=0.9\linewidth]{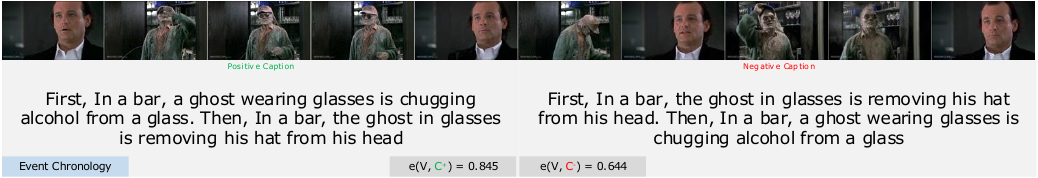}
\caption{\benchmark{} samples where OV-72B classifies the sample correctly based on the \classicvle{} criteria, but not on \strictvle{}.
In the \qabluetext{Action Binding} example, a man in a black plaid jacket is \underline{cowering}. The negative caption ($C^{-}$) changes the action from ``cowering" to ``talking aggressively." Although the model assigns a high entailment score of 0.870 to the positive caption ($C^{+}$), it also assigns a relatively high score of 0.531 to the negative caption ($C^-$).
While this satisfies the \classicvle{} criterion, it fails to meet the \strictvle{} criterion.
}
\label{fig:qa_classicvle}
\end{figure*}

\begin{figure*}[t]
\centering
\includegraphics[width=0.9\linewidth]{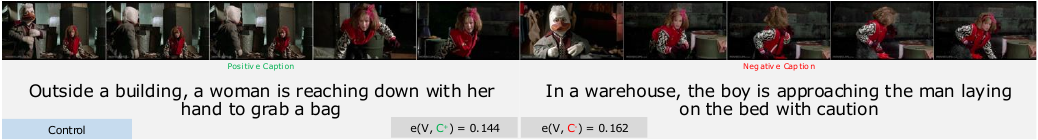}\vspace{0.3em}
\includegraphics[width=0.9\linewidth]{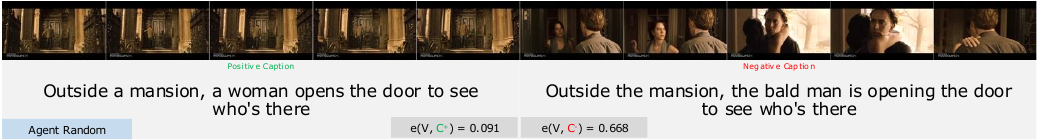}\vspace{0.3em}
\includegraphics[width=0.9\linewidth]{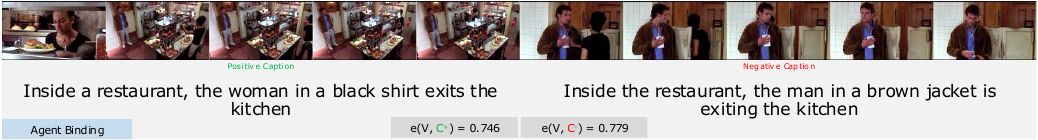}\vspace{0.3em}
\includegraphics[width=0.9\linewidth]{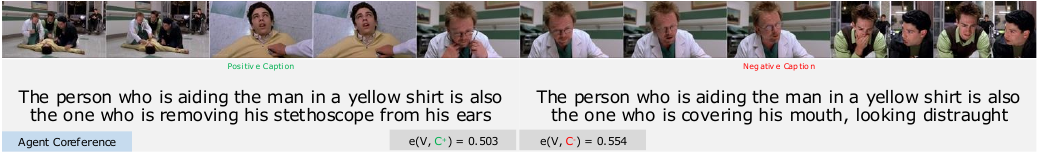}\vspace{0.3em}
\includegraphics[width=0.9\linewidth]{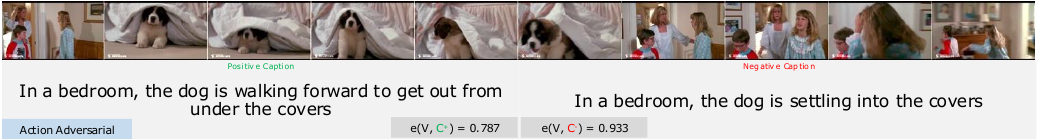}\vspace{0.3em}
\includegraphics[width=0.9\linewidth]{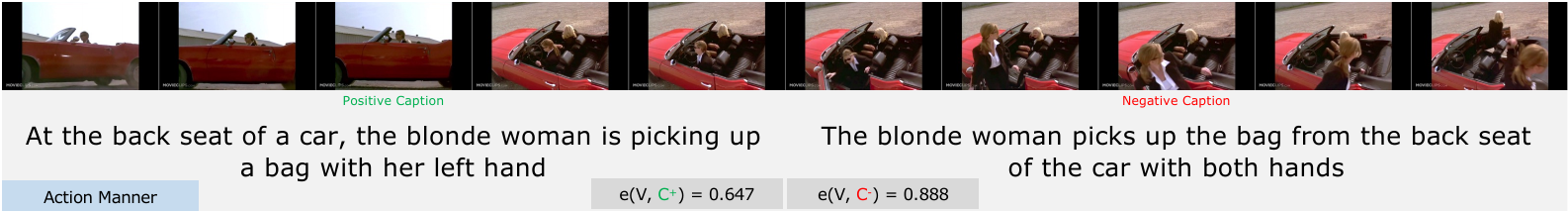}\vspace{0.3em}
\includegraphics[width=0.9\linewidth]{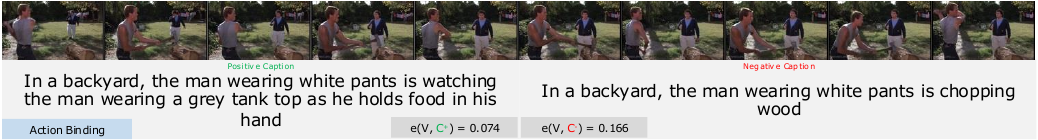}\vspace{0.3em}
\includegraphics[width=0.9\linewidth]{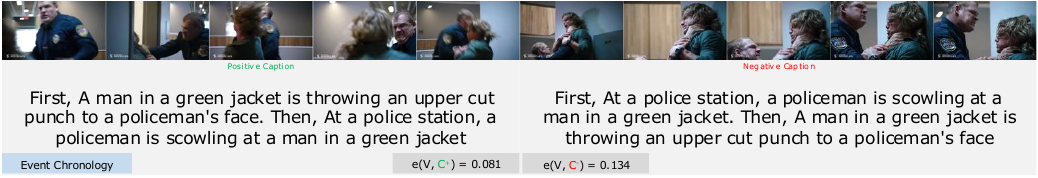}
\caption{\benchmark{} samples classified incorrectly even for \classicvle{}.
In \qabluetext{Agent Random}, the scene describes a woman opening the door for a man and hugging him. $C^{-}$ replaces the person opening the door with a random person (a bald man), and the model makes a mistake - scoring the negative caption (0.668) considerably more than the positive caption (0.091).
\qabluetext{Action Manner} has a video of two women driving into the scene where a blonde woman picks up a bag from the backseat using her left arm. The $C^{-}$ modifies how the bag is picked up - with both hands, which is clearly incorrect. However, the model makes a mistake and prefers the negative caption (0.888) over the positive caption (0.647). 
}
\label{fig:qa_incorrect}
\end{figure*}

We present examples from the OV-72B model on our benchmark for three following cases:
(i)~Samples satisfying the \strictvle{} criteria ($e(V,C^{+}) > 0.5 \land e(V,C^{-}) < 0.5$) are shown in \cref{fig:qa_strictvle};
(ii)~Samples \textit{only} satisfying the \classicvle{} condition ($e(V,C^{+}) > e(V,C^{-})$), but failing on the \strictvle{} condition are in \cref{fig:qa_classicvle}.
(iii)~Finally, samples classified incorrectly according to \classicvle{} ($e(V,C^{+}) < e(V,C^{-})$), are presented in \cref{fig:qa_incorrect}. Note these are also incorrect for \strictvle{}.
In each case, we show 10 frames from the video, the positive and negative captions, and the corresponding entailment scores.
The test name is indicated in the bottom left.

\clearpage
\section{Benchmark Creation and Details}
\label{suppsec:benchmark}

In this section, we provide details about our benchmark.
In particular, we
share all prompts used for creating positive captions and various tests (\cref{suppsubsec:srl_to_poscap}, \cref{suppsubsec:test_creation_prompts}),
share our process on creating a benchmark subset for evaluating closed models (\cref{suppsubsec:subset_selection}),
provide benchmark statistics (\cref{suppsubsec:benchmark_stats}),
discuss the strategy used to manually verify and clean all the tests (\cref{suppsubsec:quality_control}),
and finally
provide some compute and runtime details that are required to evaluate on our benchmark (\cref{suppsubsec:compute_details}).

\subsection{Prompt for Converting SRL Dictionary to a Positive Caption}
\label{suppsubsec:srl_to_poscap}

The prompt for generating the positive caption given an SRL dictionary is shown in \cref{fig:poscap_prompt}.
This refers to the discussion from Sec. 3.1 in the main paper.
We use a two-stage strategy that first inserts all elements of the SRL dictionary in a sentence and then refines it for proper grammatical structure.
\begin{figure}[b]
\centering
\noindent\begin{minipage}{1.0\textwidth}
\mdfsetup{%
middlelinewidth=1pt,
backgroundcolor=cyan!10, 
innerleftmargin=0.5cm,
innerrightmargin=0.5cm,
font=\small,
roundcorner=15pt}
\begin{mdframed}
\vspace{0.2em}
\textbf{System Prompt}

Using the provided dictionary containing verb and argument-role pairs in the style of PropBank, follow these steps to generate two captions\\

Naive Caption: Generate a caption that faithfully reflects all details from the dictionary without adding or omitting any information. 
Ensure that every argument detail is accurately included in the Naive Caption. \\

Fluent Caption: If the Naive Caption is already fluent and naturally phrased, directly copy it to the Fluent Caption. If necessary, refine the Naive Caption for improved language
fluency while strictly maintaining all original details and arguments from the dictionary. \\

Please proceed with generating the Naive Caption first, ensuring it remains comprehensive and accurate based on the provided dictionary entries.
Then, if adjustments are needed to enhance fluency, refine the Naive Caption into the Fluent Caption while ensuring that no details are overlooked or omitted.

\vspace{3mm}
\hrule
\vspace{3mm}

\textbf{Few Shot Example 1}
\begin{verbatim}
{'Verb':'walk (walk)',
'Arg0 (walker)':'man in suit',
'ArgM (direction)':'into room',
'ArgM (manner)':'slowly',
'Scene of the Event':'Warehouse'}
\end{verbatim}

Naive Caption: In a warehouse, a man in suit is walking slowly into the room.\\
Fluent Caption: In a warehouse, a man in suit is walking slowly into the room.\\

\hrule
\vspace{3mm}
\textbf{Few Shot example 2}

\begin{verbatim}
{'Verb':'burn (cause to be on fire)',
'Arg0 (thing burning)':'Wreckage',
'ArgM (location)':'Wreckage'}
\end{verbatim}

Naive Caption: The wreckage is burning on the wreckage.\\
Fluent Caption: The wreckage is burning.
\end{mdframed}
\end{minipage}
\caption{Prompt to generate the positive caption given an SRL dictionary.}
\label{fig:poscap_prompt}
\end{figure}

\subsection{Prompts for Creating Test Samples}
\label{suppsubsec:test_creation_prompts}

The prompt above (\cref{fig:poscap_prompt}) helps create the positive caption for multiple tests.
Specifically, \argen{}, \arghn{}, \verbgen{}, \manner{}, and \verbret{}, all use the above strategy, while \coref{} and \order{} adopt templates that are filled in with the complete (or partial) positive captions.

The negative prompts for \argen{}, \arghn{}, and \verbret{} are also created in the same manner as above by first replacing the specific \texttt{Verb} or \texttt{Arg0} in the dictionary followed by strategy above.

Finally, the prompt for generating the \verbgen{} negative caption is shown in \cref{fig:actadv_prompt} and for \manner{} negative captions in \cref{fig:actmod_prompt}.
Both involve generating a \texttt{target} replacement that seems reasonable followed by converting the SRL dictionary into a caption.

\begin{figure}[t]
\centering
\noindent\begin{minipage}{\textwidth}
\mdfsetup{%
middlelinewidth=1pt,
backgroundcolor=cyan!10,
innerleftmargin=0.5cm,
innerrightmargin=0.5cm,
font=\small,
roundcorner=15pt}
\begin{mdframed}
\vspace{0.2em}

\textbf{System Prompt}\\
Your objective is to generate a contradiction caption using the provided PropBank style ``input dictionary" and the `Verb' labelled as `source' based on a specific ``misalignment scenario" called ``verb misalignment". In this scenario, you should suggest an alternative contradictory value for the ``source" and label it as ``target".\\

\textbf{Key Requirements}
\begin{enumerate}
    \item ``naive caption + verb misalignment": should be plausible and could theoretically occur in real life.
    \item The ``fluent caption + verb misalignment": If the ``naive caption + verb misalignment" is already fluent and naturally phrased, directly copy it to the ``fluent caption + verb misalignment". If necessary, refine the ``naive caption + verb misalignment" for improved language fluency while strictly maintaining all original details and arguments from the dictionary \\
\end{enumerate}

\textbf{Guidelines}
\begin{enumerate}
    \item The ``target" should introduce a contradiction when compared to ``source", without being a mere negation.
    \item The ``naive caption + verb misalignment" should be clearly distinguishable from the scene described by the ``input dictionary" and should be visually distinguishable.
    \item Your replacements should be creative yet reasonable.
    \item If adjustments are needed to enhance fluency, refine the ``naive caption + verb misalignment" into the ``fluent caption + verb misalignment" while ensuring that no details are overlooked or omitted. \\
\end{enumerate}

\vspace{3mm}
\hrule
\vspace{3mm}

\textbf{Few Shot Example 1}
\begin{verbatim}
{'Verb': 'speak (speak)'},
'Arg0 (talker)': 'a man with dark hair',
'Arg2 (hearer)': 'old man'
'ArgM (manner)': 'greeting him',
'Scene of the Event': 'warehouse'}
\end{verbatim}

Target: Ignore\\
Naive Caption: On the front porch, a man with dark hair is ignoring an old man, greeting him.\\
Fluent Caption: On the front porch, a man with dark hair is ignoring an old man.

\vspace{3mm}
\hrule
\vspace{3mm}

\textbf{Few Shot Example 2}
\begin{verbatim}
{'Verb': 'open (open)',
'Arg0 (opener)': 'woman with long hair',
'Arg1 (thing opening)': 'the front door',
'ArgM (manner)': 'slowly',
'Scene of the Event': 'inside a house'}
\end{verbatim}
Target: Close\\
Naive Caption: Inside a house, a woman with long hair is closing the front door slowly.\\
Fluent Caption: Inside a house, a woman with long hair is closing the front door slowly.\\

\end{mdframed}
\end{minipage}
\caption{Prompt to generate the negative caption for \verbgen{}.}
\label{fig:actadv_prompt}
\vspace{-1em}
\end{figure}

\begin{figure}[t]
\centering
\noindent\begin{minipage}{\textwidth}
\mdfsetup{%
middlelinewidth=1pt,
backgroundcolor=cyan!10,
innerleftmargin=0.5cm,
innerrightmargin=0.5cm,
font=\small,
roundcorner=15pt}
\begin{mdframed}
\vspace{0.2em}

\textbf{System Prompt}\\
Your objective is to generate a contradiction caption using the provided PropBank style ``input dictionary" and the `ArgM (manner)' labeled as `source' based on a specific ``misalignment scenario" called ``manner misalignment". In this scenario, you should suggest an alternative contradictory value for the ``source" and label it as ``target" \\

\textbf{Key Requirements}
\begin{enumerate}
    \item ``naive caption + manner misalignment": should be plausible and could theoretically occur in real life.
    \item The ``fluent caption + manner misalignment": If the ``naive caption + manner misalignment" is already fluent and naturally phrased, directly copy it to the ``fluent caption + manner misalignment". If necessary, refine the ``naive caption + manner misalignment" for improved language fluency while strictly maintaining all original details and arguments from the dictionary.\\
\end{enumerate}

\textbf{Guidelines}
\begin{enumerate}
    \item The ``target" should introduce a contradiction when compared to ``source", without being a mere negation.
    \item The ``naive caption + manner misalignment" should be clearly distinguishable from the scene described by the ``input dictionary."
    \item Your replacements should be creative yet reasonable.
    \item If adjustments are needed to enhance fluency, refine the ``naive caption + manner misalignment" into the ``fluent caption + manner misalignment" while ensuring that no details are overlooked or omitted
\end{enumerate}

\vspace{3mm}
\hrule
\vspace{3mm}

\textbf{Few Shot Example 1}
\begin{verbatim}
{'Verb': 'look (vision)',
'Arg0 (looker)': 'a man wearing all black',
'Arg1 (thing looked at or for or on)': 'a building'
'ArgM (direction)': 'infront of him',
'ArgM (manner)': 'breathing heavily',
'Scene of the Event': 'warehouse'}
\end{verbatim}

Target: \textit{Whistling}

Naive Caption: Outside, a man wearing all black is looking in front of him at a building while whistling.

Fluent Caption: Outside, a man wearing all black is looking at a building in front of him while whistling.

\vspace{3mm}
\hrule
\vspace{3mm}

\textbf{Few Shot Example 2}
\begin{verbatim}
{'Verb': 'burn (cause to be on fire)',
'Arg0 (agent, entity causing something to be suspended)': 'climbing ropes',
'Arg1 (thing suspended)': 'woman in pink shirt',
'Arg2 (suspended from)': 'climbing ropes',
'ArgM (location)': 'on the face of the rocks',
'ArgM (manner)': 'precariously'}
\end{verbatim}

Target: Securely

Naive Caption: climbing ropes are hanging the woman in a pink shirt securely on the face of the rocks.

Fluent Caption: The woman in a pink shirt is hanging on the face of the rocks from the climbing ropes securely.

\end{mdframed}
\end{minipage}
\caption{Prompt to generate the negative caption for \manner{}.}
\label{fig:actmod_prompt}
\vspace{-1em}
\end{figure}

\subsection{Subset Creation}
\label{suppsubsec:subset_selection}
We created a subset of \benchmark{} with 150 samples in each test.
The subset was curated through random tries such that the \strictvle{} performance of the OV-72B model was comparable to the full set, allowing for fair comparisons.

\subsection{Benchmark Statistics}
\label{suppsubsec:benchmark_stats}

We present some statistics highlighting the diversity and nuance in the \benchmark{} benchmark.
Since
this benchmark is a subset of VidSitu~\cite{vidsitu}, we observe similar trends as presented in their work.

\begin{figure*}
\centering
\includegraphics[width=\linewidth]{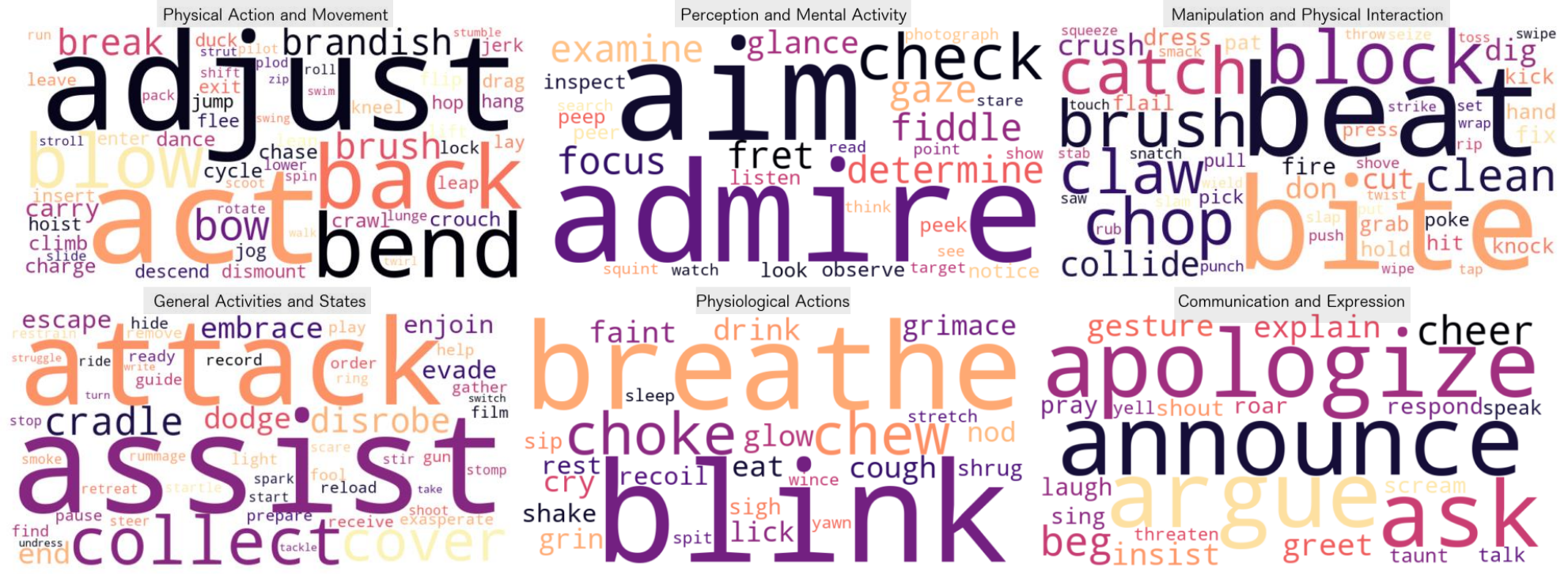}
\caption{Word-cloud of some actions in \benchmark{} in different action categories as suggested by Gemini-1.5 Pro.
Word \textbf{size does not correspond to frequency} and is assigned randomly for visualization.}
\label{fig:wc}
\end{figure*}

Videos in our benchmark are complex as there are multiple agents performing various
actions.
Actions in \benchmark{} are fine-grained.
We analyze the set using Gemini-1.5-Pro which broadly categorizes actions into 6 groups: 
physical action and movement,
communication and expression,
manipulation and physical interaction, 
perception and mental activity,
physiological actions, and
general activities and states.
In general, models struggle slightly more with physiological actions (performance $\sim$10\% lower) as compared to the average.
Some verbs from these categories are shown  in~\cref{fig:wc}, note that the size of the word here does \emph{not} correspond to its frequency in the dataset.

\begin{figure*}
\centering
\includegraphics[width=\textwidth]{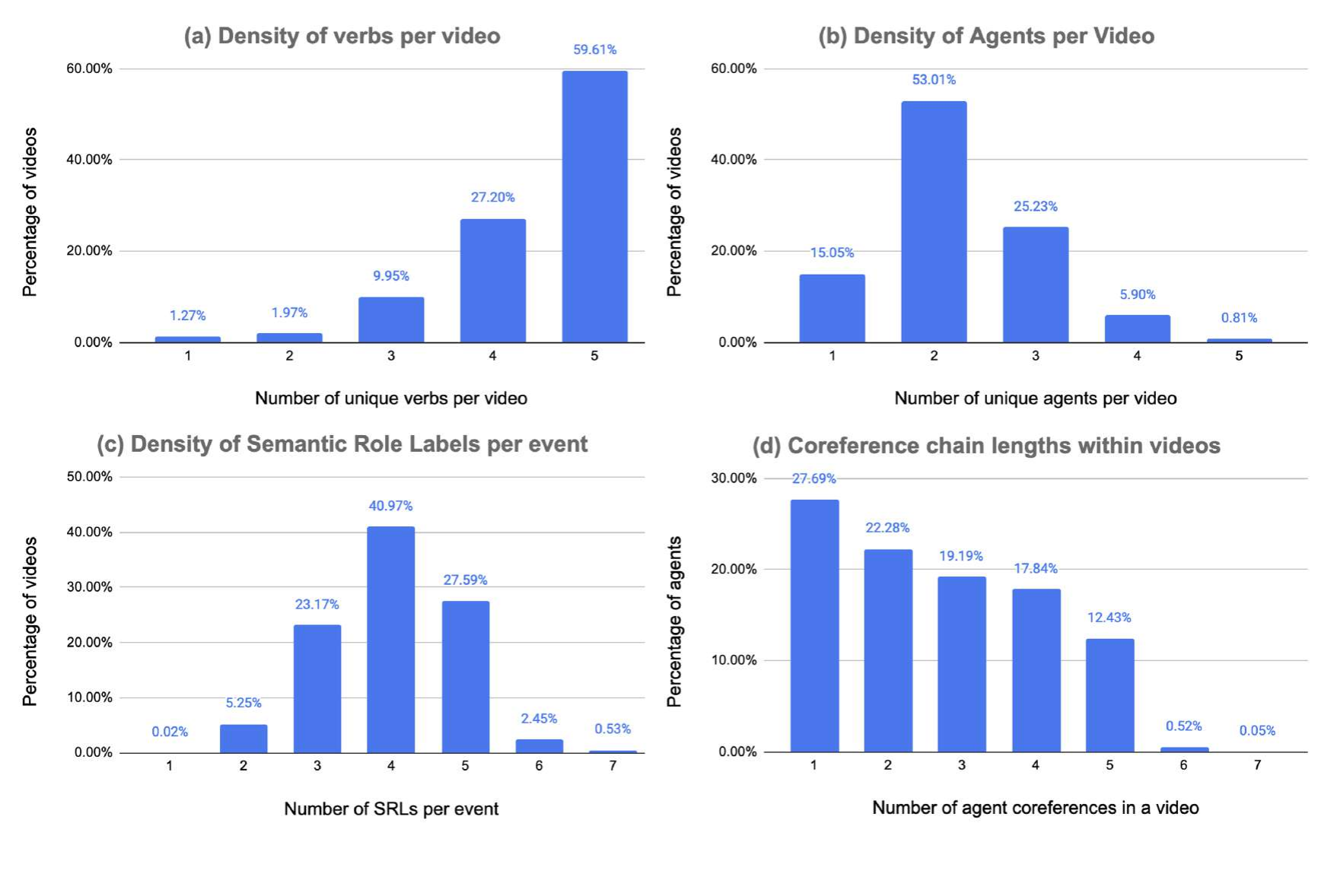}
\vspace{-10mm}
\caption{Statistics of various features of the \benchmark{} benchmark.
(a) and (b) show the distribution of verbs and agents per video, respectively.
(c) shows the density of SRL annotations per event; and
(d) shows the distribution of agent coreference lengths.
Even with short videos, the complexity of the VidSitu annotations make the task challenging.
}
\label{fig:assorted_eda}
\end{figure*}

\cref{fig:assorted_eda}a shows that around 87\% of the videos contain 4 or more unique verbs, and \cref{fig:assorted_eda}b shows that about 85\% of videos contain 2 or more unique agents (people performing actions).
We evaluate binding by leveraging the fact that one agent can perform multiple actions in the video, and the richness of the SRL annotations ensure that these events are described adequately.
In \cref{fig:assorted_eda}c, we observe that over 70\% of the events contain 4 or more SRLs (\eg~agent, patient, manner, \etc.), indicating the detail-oriented nature of the annotations. 
Finally, \cref{fig:assorted_eda}d shows that over 72\% of agents occur twice or more in their corresponding video annotation.
These agents would likely be performing two different actions, and we utilize this to create two references to the same agent in tests such as \coref{}.

\subsection{Quality Control}
\label{suppsubsec:quality_control}
\begin{wraptable}{r}{0.3\linewidth}
\centering
\tabcolsep=0.12cm
\vspace{-6mm}
\begin{tabular}{c c c c}
\toprule
Test & Videos & \# Samples & Subset \\
\midrule
Ctrl         & 850    & 2635    & 150    \\
\midrule
\argensh{}   & 588    & 873    & 150    \\
\arghnsh{}   & 615    & 1459    & 150    \\
\corefsh{}   & 183    & 339     & 150    \\
\verbgensh{} & 355    & 438     & 150    \\
\mannersh{}  & 378    & 458     & 150    \\
\verbretsh{} & 540    & 1356    & 150    \\
\ordersh{}   & 521    & 1234    & 150    \\
\bottomrule
\end{tabular}
\caption{Number of videos and samples across different tests in \benchmark.}
\label{tab:quality_control}
\end{wraptable}

To ensure that the data generated from the automated pipelines discussed earlier are correct, we filtered the data samples manually, following specific guidelines discussed in this section.
The final count of the data samples is reported in \cref{tab:quality_control}.

\paragraph{SRL dictionary to caption.}
The instructions and the interface for evaluating caption quality is described in \cref{fig:check_poscap}.
For each sample, three choices were provided: positive if the caption is correct,
negative if the caption is wrong, and
neutral if the caption cannot be negative but contains some ambiguity due to which it could not be considered positive.
Out of the 380 samples that were manually verified, 356 were marked as positive, 21 were neutral, and 3 were negative.
The number of positive and neutral samples was high (99.2\%).

\paragraph{All tests.}
For each sample of all tests, we perform a meticulous cleanup.
The instructions and the interface are presented in \cref{fig:data_cleaning}.
For each video, the green bar contains a positive caption, and the red bar contains a negative caption.
Unlike human evaluations, the positive and the negative captions are known while filtering.
Only the samples for which both positive and negative captions are deemed appropriate are retained.

\subsection{Runtime and Compute Details}
\label{suppsubsec:compute_details}
While benchmarks on long videos are interesting~\cite{videomme, mmbenchvideo}, \benchmark{} proposes important challenges that every \vllm{} needs to solve.
The short \SI{10}{\second} videos enable fast evaluation and make the benchmark accessible:
running OV-7B on all tests (except the \control{}) takes about 2.6 hours on a single RTX 4090 GPU (24 GB).
\section{Model Evaluation Prompts}
\label{suppsec:model}

We present the prompts used for all open \vllm{}s, \geminiflash{}, and \gpt{}.
The entailment and MC evaluation prompts for open models, such as \qwenvl{}, \ov{}, and \geminiflash{} are provided in~\cref{fig:vllm_entail_mc_prompt}.
Prompts for \gpt{} are shown in~\cref{fig:gpt_entail_mc_prompt}.
Note that \gpt{} is provided the explicit instruction of being provided frames of a video, while others are directly given a video.

Although some closed models have started optionally sharing logits, they are restricted to a limited top-K set,
\eg~top-20 for \gpt{}.
Hence, the logits for the \quoteyes{} and \quoteno{} tokens may not always be included in these top-k values.
To ensure the evaluation of closed models covers maximum data samples, the prompts were slightly modified to explicitly include the instruction: ``Just answer with either Yes or No.".

\begin{figure}[h]
\centering
\noindent\begin{minipage}{1.0\textwidth}
\mdfsetup{%
middlelinewidth=1pt,
backgroundcolor=cyan!10,
innerleftmargin=0.5cm,
innerrightmargin=0.5cm,
font=\small,
roundcorner=15pt}
\begin{mdframed}
\vspace{0.2em}

\textbf{Entailment Prompt}\\
Carefully watch the video and pay attention to the sequence of events, the details and actions of persons.\\
Here is a caption that describes the video: ${Caption}$ \\
Based on your observation, does the given video entail the caption?
\end{mdframed}
\end{minipage}

\noindent\begin{minipage}{1.0\textwidth}
\mdfsetup{%
middlelinewidth=1pt,
backgroundcolor=cyan!10,
innerleftmargin=0.5cm,
innerrightmargin=0.5cm,
font=\small,
roundcorner=15pt}
\begin{mdframed}
\vspace{0.2em}

\textbf{MC Prompt} \\
Carefully watch the video and pay attention to the sequence of events, the details and actions of persons.\\
Here are two captions that describe the video. \\
A) ${Caption_{1}}$\\
B) ${Caption_{2}}$\\
Based on your observation, select the caption that best describes the video.\\
Just print either A or B.

\end{mdframed}
\end{minipage}
\vspace{-2mm}
\caption{Prompts for \textbf{Open \vllm{}s} and \textbf{\geminiflash{}} and \textbf{Gemini-1.5-Pro}.
\textbf{Top}: Entailment evaluation prompt.
\textbf{Bottom}: Multiple-choice evaluation prompt.}
\label{fig:vllm_entail_mc_prompt}
\end{figure}

\begin{figure}[ht]
\centering
\noindent\begin{minipage}{1.0\textwidth}
\mdfsetup{%
middlelinewidth=1pt,
backgroundcolor=cyan!10,
innerleftmargin=0.5cm,
innerrightmargin=0.5cm,
font=\small,
roundcorner=15pt}
\begin{mdframed}
\vspace{0.2em}

\textbf{Entailment Prompt} \\
You are given frames sampled sequentially from a video. Carefully watch the video frames and pay attention to the sequence of events, the details and actions of persons.\\
Here is a caption that describes the video: ${Caption}$\\
Based on your observation, does the given video entail the caption?\\
Just answer with either Yes or No.

\end{mdframed}
\end{minipage}
\vspace{0.2em}
\noindent\begin{minipage}{1.0\textwidth}
\mdfsetup{%
middlelinewidth=1pt,
backgroundcolor=cyan!10,
innerleftmargin=0.5cm,
innerrightmargin=0.5cm,
font=\small,
roundcorner=15pt}
\begin{mdframed}
\vspace{0.2em}

\textbf{MC Prompt} \\
You are given frames sampled sequentially from a video. Carefully watch the video frames and pay attention to the sequence of events, the details and actions of persons.\\
Here are two captions that describe the video.\\
A) ${Caption_{1}}$ \\
B) ${Caption_{2}}$ \\
Based on your observation, select the caption that best describes the video.\\
Just print either A or B.

\end{mdframed}
\end{minipage}
\vspace{-2mm}
\caption{Prompts for \textbf{\gpt}.
\textbf{Top}: Entailment evaluation prompt.
\textbf{Bottom}: Multiple-choice evaluation prompt.}\label{fig:gpt_entail_mc_prompt}
\end{figure}

\section{Limitations}
\label{suppsec:limitations}

We discuss some limitations of our work.

\begin{enumerate}
\item One of the shortcomings is the limited ability to scale the benchmark.
\benchmark{} relies on SRLs, which are obtained from careful (and costly) human annotations~\cite{vidsitu}.
Further, we use \llm{}s to generate captions from the SRL dictionary and to create several tests (\cref{suppsubsec:srl_to_poscap}, \cref{suppsubsec:test_creation_prompts}).
However, \llm{}s are prone to hallucinations, and hence, we do a round of human verification to confirm that the captions are appropriate.
Thus, costly human intervention is required from SRL curation to verification of individual test samples.

\item \benchmark{} is not intended as a one-stop benchmark to evaluate all abilities of \vllm{}s.
Instead, it evaluates \vllm{}s for facets of compositionality, a fundamental aspect of visio-linguistic reasoning.
Also, as \benchmark{} is derived from VidSitu, a person-centric dataset, our benchmark focuses on people and their actions/interactions.

\item Lastly, our proposed \strictvle{} metric cannot be used to evaluate contrastive models, as these models do not provide a direct \quoteyes{} probability.
When the alignment score is used as a proxy to the entailment score (similar to~\cite{vitatecs}), we show that contrastive CLIP-based models do not perform well even with \classicvle{} and are therefore unlikely to be competitive at a stricter entailment.
\end{enumerate}

\begin{figure*}[p]
\centering
\includegraphics[width=0.80\textwidth]{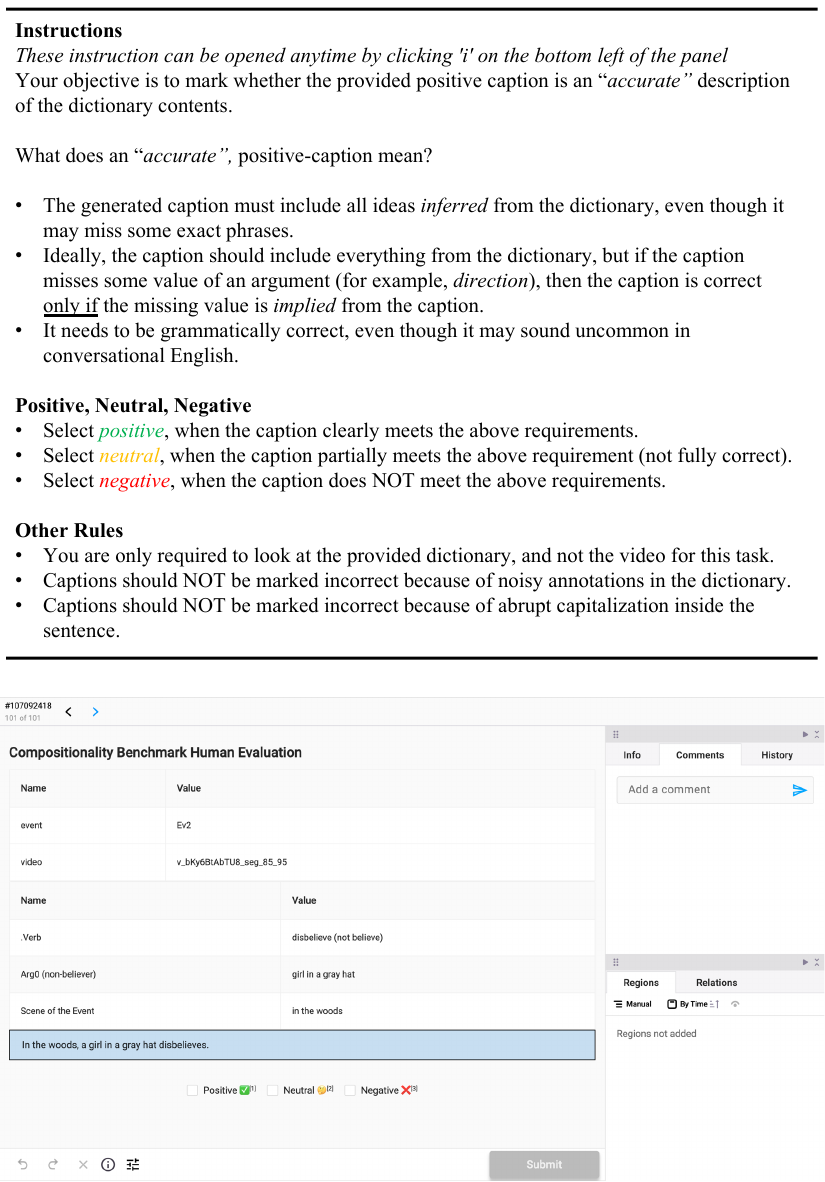}
\caption{Instructions and interface to verify the quality of captions generated from LLaMA-3-70B.}
\label{fig:check_poscap}
\end{figure*}

\begin{figure*}[p]
\centering
\includegraphics[width=0.80\textwidth]{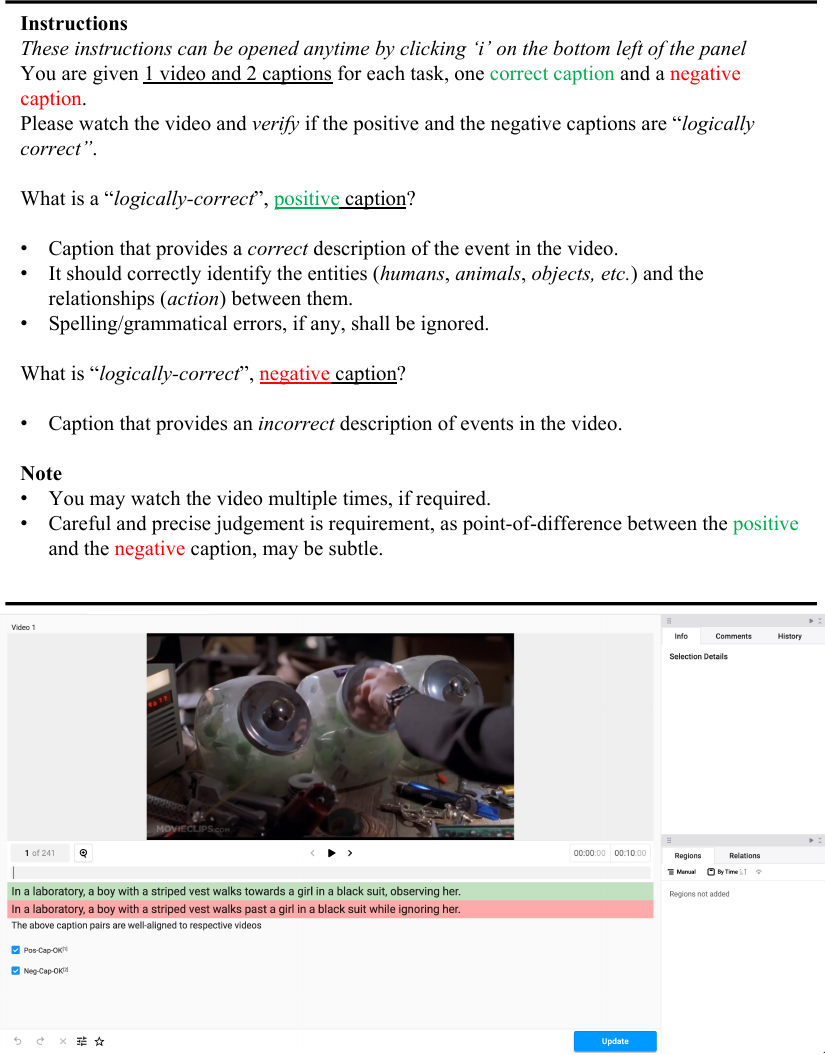}
\caption{Data cleaning instruction for all the tests.}
\label{fig:data_cleaning}
\end{figure*}

\end{document}